\documentclass[letterpaper, 10 pt, conference]{ieeeconf}  

\IEEEoverridecommandlockouts                              

\usepackage{graphicx}
\usepackage{booktabs}
\usepackage{makecell}
\usepackage{subcaption}
\usepackage{siunitx}
\usepackage{amsmath}
\usepackage{multirow}
\usepackage{threeparttable}
\usepackage{hyperref}

\title{\LARGE \bf
Antagonistic Bowden-Cable Actuation of a Lightweight Robotic Hand: Toward Dexterous Manipulation for Payload-Constrained Humanoids
}

\author{Sungjae Min$^{1}$, Hyungjoo Kim$^1$ and David Hyunchul Shim$^{1}$
\thanks{\textit{(Corresponding author: Sungjae Min)}}
\thanks{$^{1}$Sungjae Min, Hyungjoo Kim and D. Hyunchul Shim are with School of Electrical Engineering,
         Korea Advanced Institute of Science and Technology, 291 Daehak-ro Daejeon, Republic of Korea
         {\tt\small \{sungjae\_min, hyungjoo\_kim, dhcshim\}@kaist.ac.kr}}%
}

\begin{document}

\maketitle
\thispagestyle{empty}
\pagestyle{empty}

\begin{abstract}

Humanoid robots toward human-level dexterity require robotic hands capable of simultaneously providing high grasping force, rapid actuation speeds, multiple degrees of freedom, and lightweight structures within human-like size constraints. Meeting these conflicting requirements remains challenging, as satisfying this combination typically necessitates heavier actuators and bulkier transmission systems, significantly restricting the payload capacity of robot arms. In this letter, we present a lightweight anthropomorphic hand actuated by Bowden cables, which uniquely combines rolling-contact joint optimization with antagonistic cable actuation, enabling single-motor-per-joint control with negligible cable-length deviation. By relocating the actuator module to the torso, the design substantially reduces distal mass while maintaining anthropomorphic scale and dexterity. Additionally, this antagonistic cable actuation eliminates the need for synchronization between motors. Using the proposed methods, the hand assembly with a distal mass of $236~\mathrm{g}$ (excluding remote actuators and Bowden sheaths) demonstrated reliable execution of dexterous tasks, exceeding $18~\mathrm{N}$ fingertip force and lifting payloads over one hundred times its own mass. Furthermore, robustness was validated through Cutkosky taxonomy grasps and trajectory consistency under perturbed actuator–hand transformations.

\end{abstract}

\section{INTRODUCTION}

The advent of humanoid robots highlights the hand as one of the most challenging yet indispensable components. To move beyond simple grasp-and-release manipulation and execute complex tasks in human environments, humanoid robots require hands with dexterity approaching that of humans. The human hand provides more than 20 degrees of freedom (DOFs) while simultaneously delivering grasping forces exceeding $400~\mathrm{N}$. It further exhibits high actuation speed, inherent backdrivability, and remarkable payload capacity, yet weighs only about $0.4~\mathrm{kg}$~\cite{estimation}. These characteristics enable dynamic movements of the whole arm without significant mass penalty and arise from the intricate arrangement of tendons and ligaments together with complex biomechanics of the soft tissues.

Reproducing these features mechanically entails a series of inherent trade-offs. Relevant factors include overall size, weight, DOFs, grasp adaptability, impact robustness, repeatability, sensor integration, ease of maintenance, arm compatibility, and cost. Among these, balancing weight against output force and motion capability is particularly critical. Achieving high torque, speed, and dexterity typically requires numerous powerful actuators, but an overly heavy hand reduces the effective payload of the arm and narrows its task applicability. Likewise, the use of high gear ratios to generate sufficient joint torque amplifies reflected inertia and friction, thereby compromising backdrivability, compliance, and impact safety. Another example is the trade-off between palm volume and wrist compatibility: embedding many actuators within the palm to realize high DOFs extends the lever arm to the wrist pivot, requiring larger arm motions and exacerbating inverse-kinematic constraints, particularly in confined spaces. Ironically, such conflicts limit the range of tasks that humanoid robots equipped with dexterous hands can practically perform.

A variety of approaches have been proposed to balance these competing requirements while retaining sufficient dexterity~\cite{journey, century, mechanical}. Some designs place actuators directly at the finger joints~\cite{KITECH}, whereas others embed them in the palm and employ linkage-based transmissions~\cite{ILDA}. Cable-driven architectures have received particular attention, as they can achieve human-like dimensions, force generation, and a high number of DOFs. The Biomimetic Hand~\cite{biomimetic} reproduced the intricate tendon and ligament structure of the human hand and achieved natural motion with only nine actuators. Similarly, the Anatomically Correct Testbed (ACT) Hand~\cite{ACT} employed spectra strings to mimic tendons and ligaments, achieving very high DOFs through cable actuation. The SeoulTech Hand~\cite{PCDRH} exploited the small differential length changes of antagonistic cable pairs to realize one-motor-per-DOF actuation and high-DOF motion. CasiaHand~\cite{casia} demonstrated that tendon-driven hands with actuators integrated into the palm can achieve 15 DOFs with human-scale dimensions. When the tendon-sheath is routed into the fingers, nonlinearities such as friction and hysteresis become more pronounced, as analyzed and compensated in a tendon-sheath driven hand~\cite{sheath}. Despite various efforts, existing cable-driven hands remain constrained by distal weight and limited wrist compatibility. In particular, bulky actuator modules increase the lever arm to the wrist pivot and hinder integration with diverse robotic arms. These unresolved limitations continue to restrict their practical deployment on humanoid robots.


For humanoid robots to perform a wide range of tasks in human-centered environments, robotic hands must satisfy most of these requirements simultaneously. In practice, imitation learning platforms and large-scale demonstration datasets for Physical AI increasingly emphasize five-fingered anthropomorphic hands. In line with this trend, specialized demonstration devices have also been introduced, such as UMI for non-anthropomorphic grippers~\cite{UMI} and DexUMI for anthropomorphic five-fingered hands~\cite{DexUMI}.

\begin{figure}[tb] 
    \centering
    \includegraphics[width=\columnwidth]{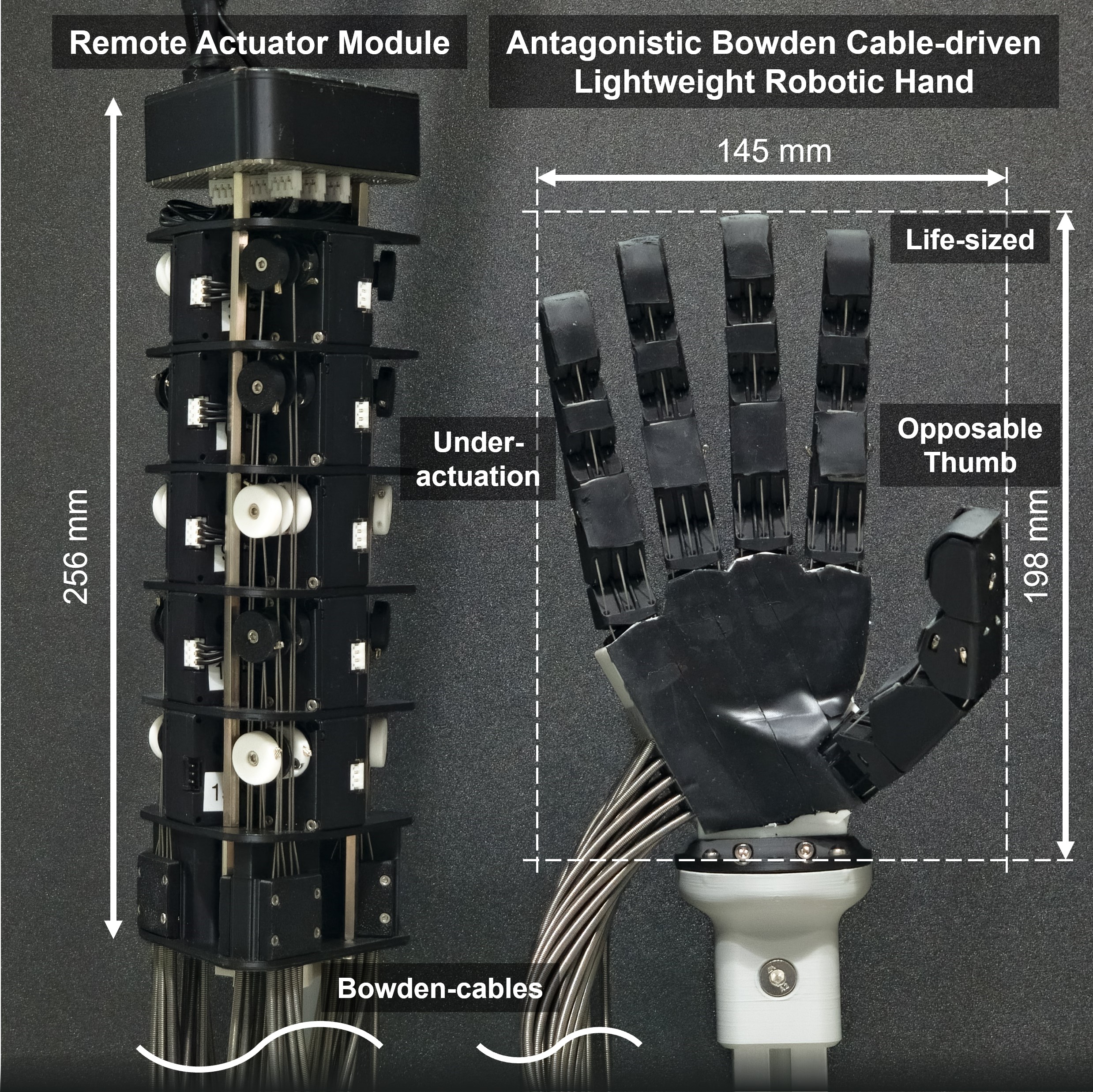}
    \caption{Overview of the proposed Antagonistic Bowden-Cable-Driven Lightweight (ABCDL) Robotic Hand.}
    \label{fig:PCDRH2}
    \vspace{-1em}
\end{figure}

\begin{table}[t]
  \caption{Specifications of the Proposed Robotic Hand}
  \label{tab:hand_characteristics}
  \centering
  \renewcommand{\arraystretch}{1.15}
  \begin{tabular}{l c}
    \toprule
    Number of fingers         & 5 \\
    Number of actuators       & 15 \\
    Degrees of freedom        & 20 \\
    \midrule
    Payload                   & 25~kg \\
    Weight (hand only)        & 236~g \\
    Weight (including $0.9~\textrm{m}$ Bowden cables) & 1262~g \\
    Dimensions (anatomical posture) & $122 \times 189 \times 198~\si{mm^3}$ \\
    Dimensions (adducted) & $56 \times 102 \times 198~\si{mm^3}$ \\
    Anthropomorphic scale     & 1:1 with human hand \\
    \bottomrule
  \end{tabular}
  \vspace{-1.5em}
\end{table}

These observations highlight the need for a lightweight, high-DOF robotic hand that resolves the trade-offs of existing designs while remaining compatible with humanoid platforms. In this letter, we propose an antagonistic Bowden-cable actuated lightweight robotic hand as a higher-level compromise among these competing requirements, as illustrated in Fig.~\ref{fig:PCDRH2}. By relocating actuators remotely (e.g., within the torso of a humanoid robot), the design achieves sufficient DOFs while minimizing distal bulk and mass. Antagonistic cable pairs driven by single actuators mitigate the unintended flexion problem of unidirectional cable hands during dorsal contact, while rolling-contact joints(RCJs) eliminate the need for complex bearings, linkages, or pins. The proposed design particularly emphasizes lightweight construction and humanoid integration, while preserving human-like dexterity through a compact modular antagonistic mechanism. The entire hand is 3D printed for low-cost fabrication, flexible cable routing, and ease of sensor integration. Parametric and modularized finger links simplify design and facilitate maintenance. The remainder of this letter is organized as follows: Section~\ref{Sec:2} describes the general specifications, including the cable-routing strategy and kinematic characteristics. Section~\ref{Sec:3} details the antagonistic Bowden-cable actuation implemented in RCJs. Section~\ref{Sec:4} presents experimental evaluations of performance, actutation consistency, and object manipulation. Section~\ref{Sec:5} addresses the concluding remarks.

\section{ANTAGONISTIC BOWDEN-CABLE DRIVEN LIGHTWEIGHT ROBOTIC HAND \label{Sec:2}}

\begin{figure}[t]
  \centering
  \begin{subfigure}{0.49\columnwidth}
    \centering
    \includegraphics[width=\linewidth]{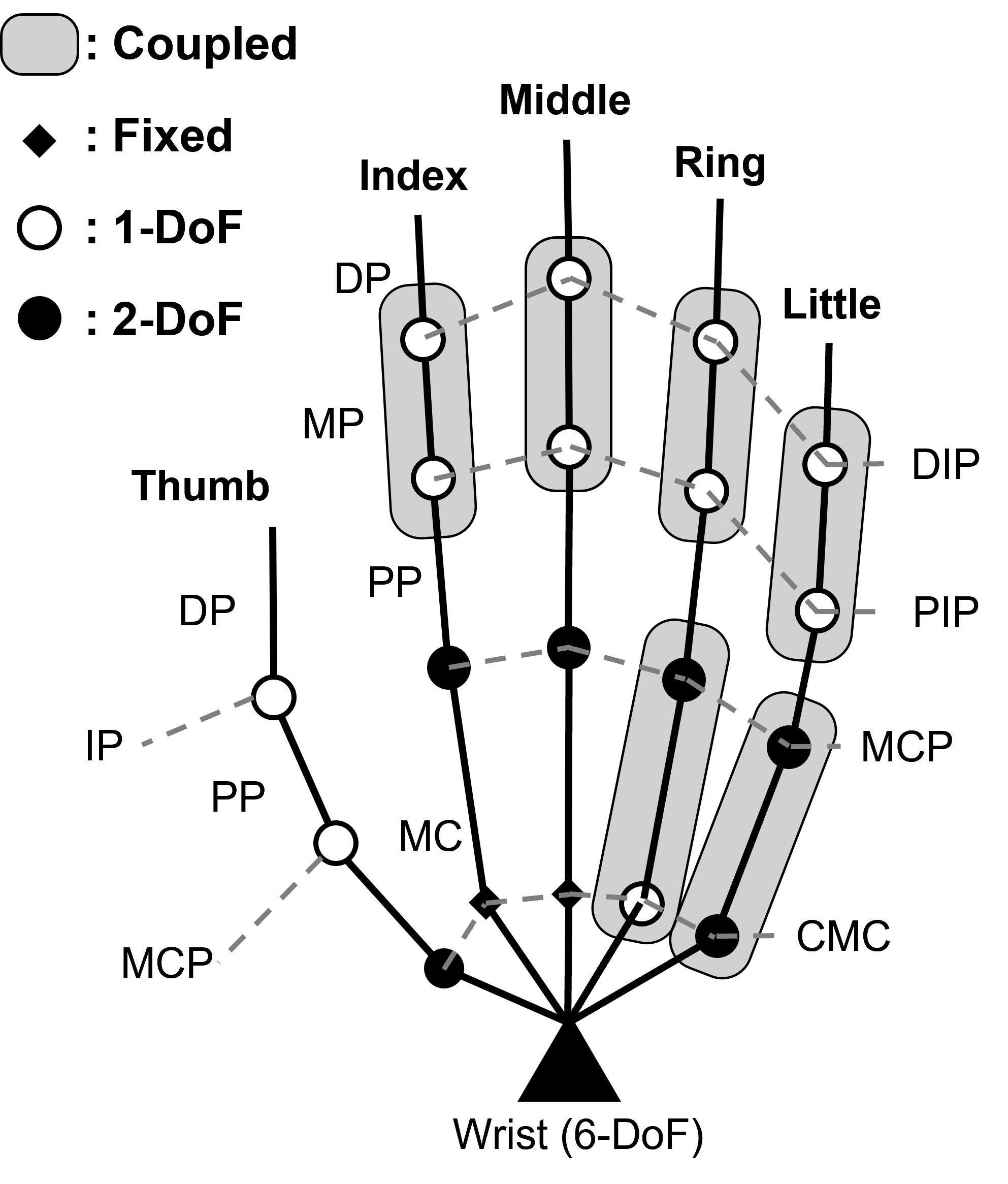}
    \caption{Human hand}
    \label{fig:human_hand}
  \end{subfigure}
  \hfill
  \begin{subfigure}{0.49\columnwidth}
    \centering
    \includegraphics[width=\linewidth]{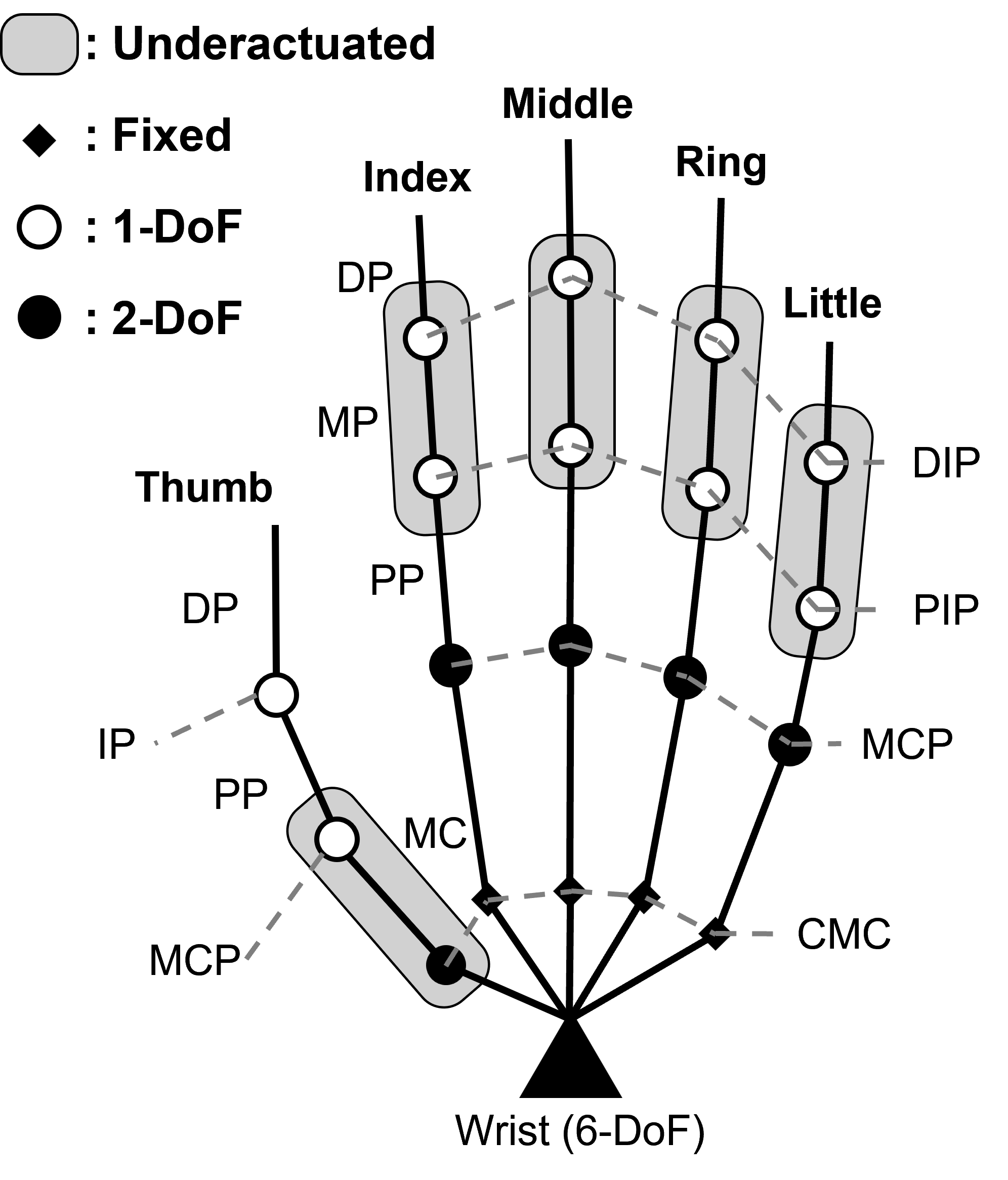}
    \caption{Proposed robotic hand}
    \label{fig:robotic_hand}
  \end{subfigure}
  \caption{Comparison of joint structures and actuation between the human hand and the proposed robotic hand.}
  \label{fig:hand_comparison}
  \vspace{-1em}
\end{figure}

\begin{table}[t]
  \caption{Ranges of Motion of Thumb and Fingers}
  \label{tab:rom_simplified}
  \centering
  \renewcommand{\arraystretch}{1.15}
  \setlength{\tabcolsep}{6pt}
  \begin{tabular}{r |c c c c}
    \toprule
    \makecell{Thumb\\Finger\\Joint} & \makecell{TM\\MCP\\flex/ext} & \makecell{TM\\MCP\\abd/add} & \makecell{MCP\\PIP\\flex/ext} & \makecell{IP\\DIP\\flex/ext} \\
    \midrule
    Thumb        & $0\sim100^\circ$ & $-45\sim45^\circ$ & $0\sim100^\circ$ & $0\sim100^\circ$ \\
    Index--Little & $0\sim100^\circ$ & $-30\sim30^\circ$ & $0\sim100^\circ$ & $0\sim100^\circ$ \\
    \bottomrule
  \end{tabular}
  \vspace{-1.25em}
\end{table}

\begin{table*}[t]
  \caption{Comparison of the Proposed Robotic Hand with Representative Robotic Hands}
  \label{tab:hand_comparison}
  \centering
  \renewcommand{\arraystretch}{1.15}
  \setlength{\tabcolsep}{3pt} 
  \begin{tabular}{l | c c c c c c c c c}
    \toprule
    Robotic hand & \makecell{Fingers\\(non-active)} & \makecell{Anthropomorphic\\scale~($\times$ human hand)} & \makecell{Weight\\(hand)~$[\mathrm{g}]$} & \makecell{Total\\weight~$[\mathrm{g}]$} & DOFs & \makecell{Number of\\Actuators} & \makecell{Fingertip\\force~$[\mathrm{N}]$} & \makecell{Under-\\actuation} & \makecell{Thumb\\opposability} \\
    \midrule
    Proposed hand & 5 & 1.0 & 236 & 1952 & 20 & 15 & 18 & Tendon & 0.172 \\
    Biomimetic Hand~\cite{biomimetic} & 5 & 1.0 & N/A & 942 & - & 10 & N/A & Compliant tissue & N/A \\
    ACT Hand~\cite{ACT} & 5(2) & 1.0 & N/A & N/A & 23 & 24 & N/A & Passive ligament & N/A \\
    SeoulTech Hand~\cite{PCDRH} & 5 & 1.0 & 191 & 934 & 20 & 15 & 6 & Tendon~/~spring & 0.049 \\
    Casia hand~\cite{casia} & 5 & 1.2 & 890 & - & 15 & 7 & 16 & Tendon~/~spring & N/A \\
    \bottomrule
  \end{tabular}
  \begin{tablenotes}
      \footnotesize
      \item \textit{Note.} `N/A': not reported in the literature. `–': not applicable to the hand's design.
    \end{tablenotes}
  \vspace{-1.5em}
\end{table*}

Fig.\ref{fig:PCDRH2} presents an overview of the proposed robotic hand system. The joints adopt a rolling-contact design actuated by antagonistic tendons, with the remote actuator module connected via Bowden-cables. Spectra threads were used to secure the RCJs, while all other components of the hand and actuator module were fabricated from polylactic acid (PLA) using a conventional fused deposition modeling (FDM) 3D printer. To compensate for the low friction coefficient of PLA, rubber tape was locally applied to the fingers and palm surfaces. The general specifications of the hand are summarized in Table~\ref{tab:hand_characteristics}. As shown, despite its relatively high number of actuated DOFs, the design matches the human hand in scale while being lighter in weight, making it well suited for payload-constrained humanoids. The actuator module employed in the experiments of Fig.\ref{fig:PCDRH2} weighs $690~\mathrm{g}$; however, owing to the remote actuation scheme, it can be placed freely on the host robot. Its configuration in terms of placement, speed, and torque can therefore be adapted to the specific requirements of the application. In addition, Table~\ref{tab:hand_comparison} provides a comparison between the proposed hand and other representative robotic hands discussed in the Introduction.

Fig.~\ref{fig:hand_comparison} compares the joint structure and degrees of freedom between the proposed robotic hand and the human hand. The human hand is generally considered to have about 23 DOFs, many of which exhibit coupled rather than independent motions. To actuate and stabilize these numerous DOFs, more than thirty intrinsic and extrinsic muscles transmit forces through tendons to the palm and forearm~\cite{handFunction, normative}.

\begin{figure}[tpb] 
    \centering
    \begin{subfigure}{0.47\columnwidth}
        \centering
        \includegraphics[width=\linewidth]{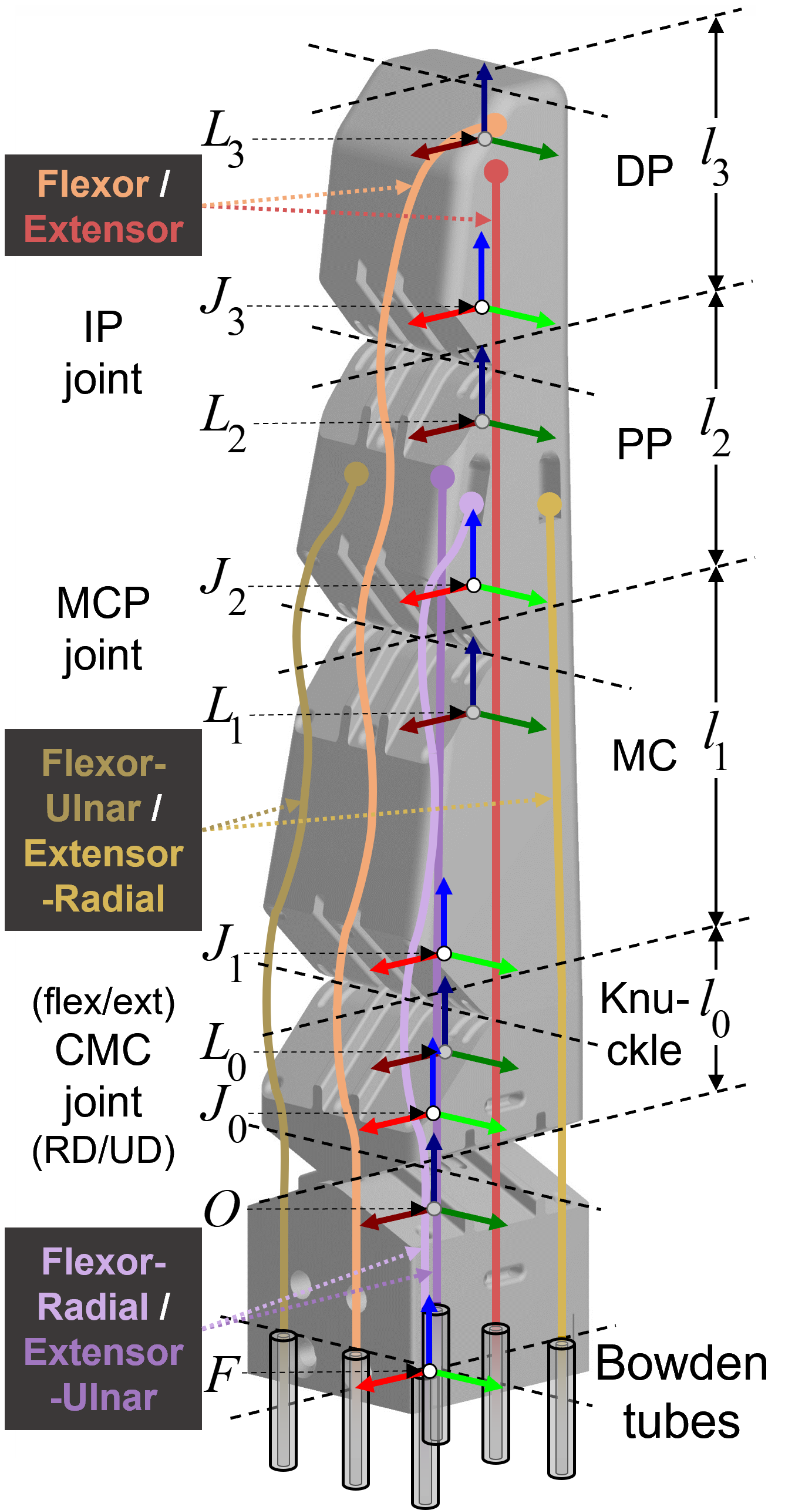}
        \caption{Thumb}
        \label{fig:ThumbStructure}
    \end{subfigure}
    \hfill
    \begin{subfigure}{0.47\columnwidth}
        \centering
        \includegraphics[width=\linewidth]{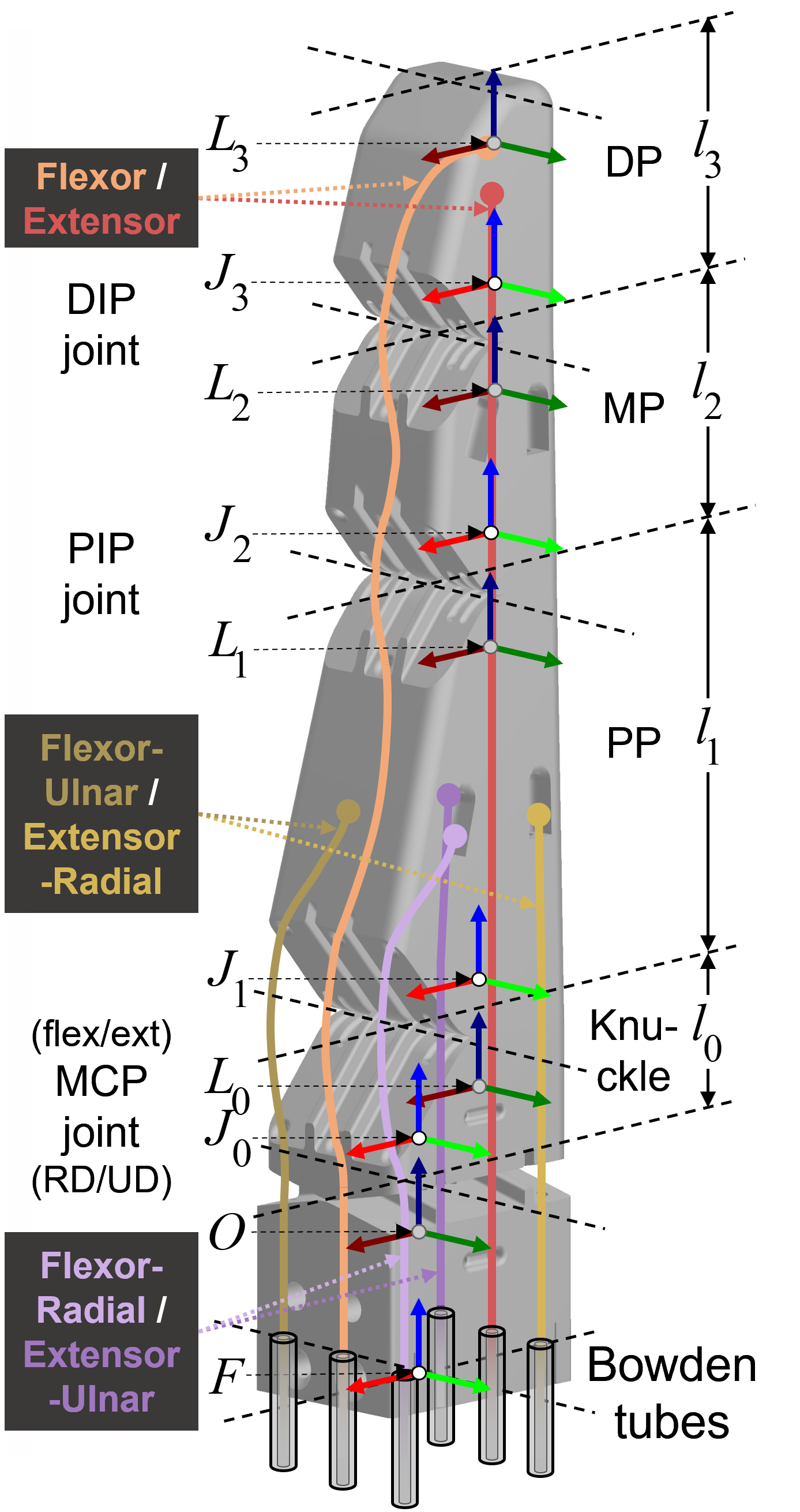}
        \caption{Index--Little}
        \label{fig:FingerStructure}
    \end{subfigure}
    \caption{Kinematic structure of the fingers in the proposed robotic hand, including cable routing and joint configuration.}
    \label{fig:FingerStructure}
    \vspace{-1.5em}
\end{figure}

The fingers of the proposed hand are designed with identical parameterized structures, except for differences in tendon routing between the thumb metacarpal (MC) and the proximal phalanges (PP) of the other fingers, as illustrated in Fig.~\ref{fig:FingerStructure}. This modular design reduces complexity in design and fabrication, while allowing the link length and width to be scaled to the desired hand size. For the four fingers, the ulnar and radial cable pairs are anchored to the PP to constrain the two degrees of freedom of the metacarpophalangeal (MCP) joint. In contrast, in the thumb the corresponding cables pass through the MC and are fixed to the PP, thereby realizing underactuation between flexion–extension of the carpometacarpal (CMC) and MCP joints. Moreover, because the cables associated with the MCP and CMC joints are routed in parallel, a single degree of freedom is effectively actuated by two tendons. This configuration enables nearly double the torque to be transmitted at these basal joints compared with a single-tendon arrangement, thereby reinforcing torque generation at the palm-side proximal joints rather than relying solely on distal fingertip forces.
The thumb's flexor–extensor cable pair independently drives the interphalangeal (IP) joint, whereas in the other fingers the flexor–extensor cables pass through the middle phalanx (MP) and are fixed at the distal phalanx (DP), thereby realizing underactuation between the proximal interphalangeal (PIP) and distal interphalangeal (DIP) joints. This design ensures that flexion torque generated at the MCP and CMC joints is sufficient to counteract the tip torque of the IP joint, so that grasping does not rely solely on fingertip contact but also engages the PP and MC links, resulting in stable and strong adaptive grasps. In addition, the cable routing inside each phalanx and knuckle was shaped along smooth curves, ensuring bending radii above the recommended minimum across the full ROM. This design consideration was intended to mitigate resistance due to cable deformation during actuation, thereby supporting tendon transmission.

\section{DESIGN OF ROBOTIC HAND \label{Sec:3}}

\subsection{Antagonistic Cable-Driven Rolling Contact Joint}

\begin{figure}[tpb] 
    \centering
    \begin{subfigure}{0.48\columnwidth}
        \centering
        \includegraphics[width=\linewidth]{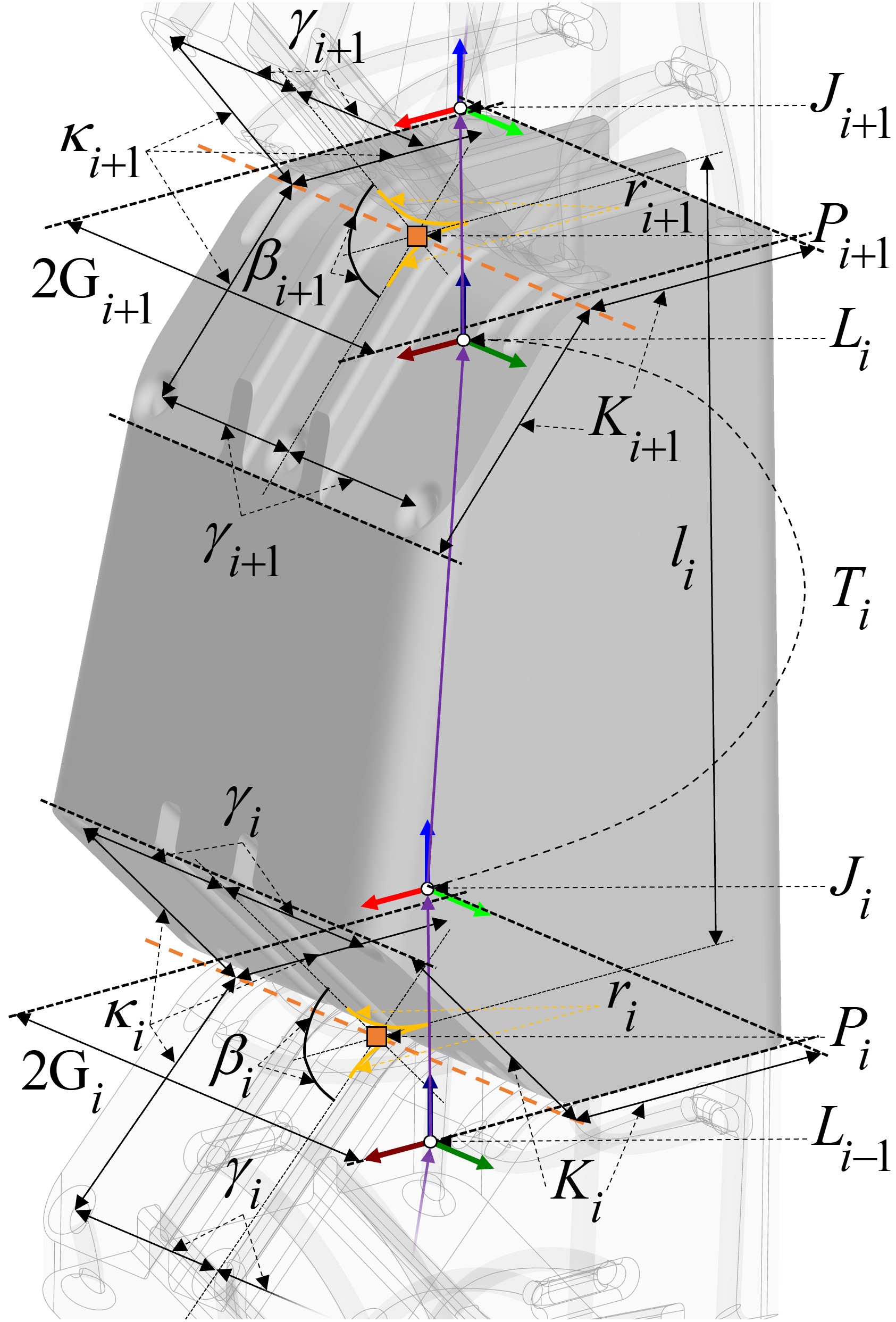}
        \caption{Phalange}
        \label{fig:RollingContactLinkKinematics}
    \end{subfigure}
    \hfill
    \begin{subfigure}{0.45\columnwidth}
        \centering
        \includegraphics[width=\linewidth]{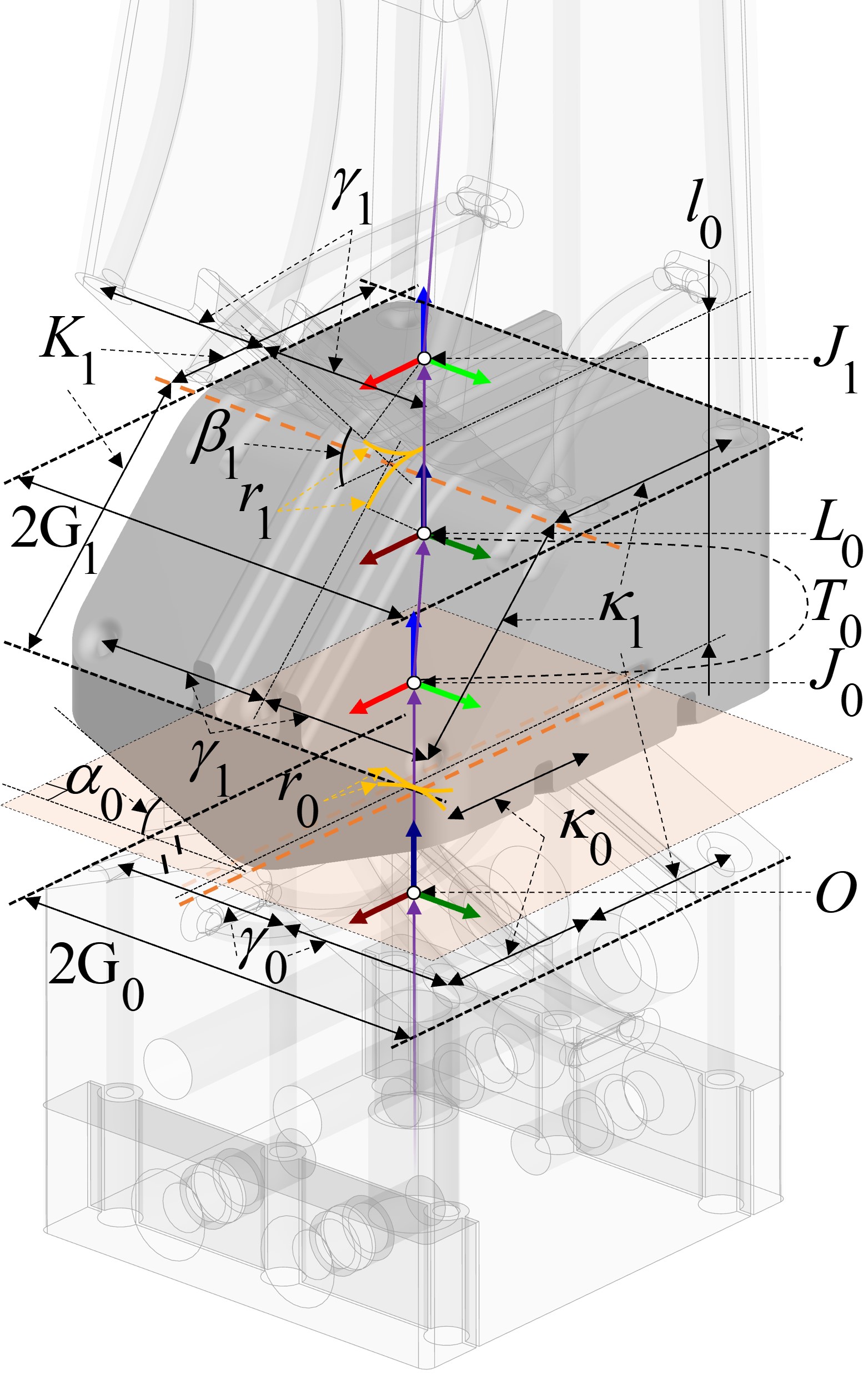}
        \caption{Knuckle}
        \label{fig:RollingContactKnuckleKinematics}
    \end{subfigure}
    \caption{Geometric definition of the joint design parameters and coordinate relationships for antagonistic cable actuation.}
    \label{fig:RollingContactKinematics}
    \vspace{-1.5em}
\end{figure}

\begin{table*}[t]
  \caption{Design Parameters of the ABCDL Hand Reflecting Anthropomorphic Scale and Joint ROM}
  \label{Tab:designedParameters}
  \centering
  \renewcommand{\arraystretch}{1.15}
  \setlength{\tabcolsep}{2.5pt}
  \begin{tabular}{r | c c c c | c c c c | c c c c | c c c c | c c c c}
    \toprule
    Parameters & \makecell{$l_0$ \\ $[\si{mm}]$} & \makecell{$l_1$ \\ $[\si{mm}]$} & \makecell{$l_2$ \\ $[\si{mm}]$} & \makecell{$l_3$ \\ $[\si{mm}]$} &
                \makecell{$\gamma_0$ \\ $[\si{mm}]$} & \makecell{$\gamma_1$ \\ $[\si{mm}]$} & \makecell{$\gamma_2$ \\ $[\si{mm}]$} & \makecell{$\gamma_3$ \\ $[\si{mm}]$} &
                \makecell{$\kappa_0$ \\ $[\si{mm}]$} & \makecell{$\kappa_1$ \\ $[\si{mm}]$} & \makecell{$\kappa_2$ \\ $[\si{mm}]$} & \makecell{$\kappa_3$ \\ $[\si{mm}]$} &
                \makecell{$\alpha_0$ \\ $[^\circ]$} & \makecell{$\beta_1$ \\ $[^\circ]$} & \makecell{$\beta_2$ \\ $[^\circ]$} & \makecell{$\beta_3$ \\ $[^\circ]$} &
                \makecell{$r_0$ \\ $[\si{mm}]$} & \makecell{$r_1$ \\ $[\si{mm}]$} & \makecell{$r_2$ \\ $[\si{mm}]$} & \makecell{$r_3$ \\ $[\si{mm}]$} \\
    \midrule
    Thumb         & 16 & 35 & 27.5 & 27.5 & 9.5 & 9.5 & 8.2 & 7.5  
                  & 9.1 & 11.7 & 10.2 & 8.7  & 22.5 & 50 & 50 & 50
                  & 3.4 & 4.5 & 3.9 & 3.3 \\
    Index--Little & 15.5 & 42.5 & 24.5 & 24.5 & 7.5 & 7.5 & 6.5 & 6
                  & 9.5 & 12.7 & 8.7 & 8.2  & 15 & 50 & 50 & 50
                  & 1.9 & 4.9 & 3.3 & 3.1 \\
    \bottomrule
  \end{tabular}
  \vspace{-1.25em}
\end{table*}

Unlike a conventional pin-pivot joint, a rolling contact joint (RCJ) generates rotation by the rolling engagement of two curved surfaces, enabling a wide range of motion without link interference or additional elements such as bearings. This operating principle resembles the motions of human anatomical joints while maintaining low friction, and therefore rolling contact joints have been actively studied in applications ranging from robot arms~\cite{AMBIDEX} to robotic hands~\cite{arcRCJ}, continuum manipulators~\cite{sigmaRCJ}, and articulated surgical endoscopes~\cite{asymRCJ, invasiveRCJ}. In particular, the shape of the contact surfaces can be designed to tune the motion trajectory and force-transmission characteristics~\cite{asymRCJ}, allowing task-specific optimization.

In this work, we employ RCJs to achieve strong, anthropomorphic finger motions while driving an antagonistic cable pair with a single motor. The contact geometry is optimized so that the combined length change of the cable pair remains negligible over the full ROM, allowing one motor to actively constrain flexion and extension without requiring two actuators. 

All finger phalanges and the thumb MC links ($i \geq 1$) adopt the octagonal base shape shown in Fig.~\ref{fig:RollingContactKinematics}(\subref{fig:RollingContactLinkKinematics}). The length of each phalanx is defined as $l_i$, with a chamfered taper toward the distal side such that $G_{i+1} \leq G_i$ and $K_{i+1} \leq K_{i}$. Rolling surfaces are arranged symmetrically about a virtual pivot $P_i$, with flexion–extension cable holes at distance $\kappa_i$ and radial–ulnar holes at spacing $\gamma_i$, which together constitute the primary moment arm of each RCJ and therefore directly influence torque and speed capacity. They were designed to be as large as possible while maintaining anthropomorphic scale. The angle $\beta_i$ between the two rolling surfaces corresponds to half the flexion–extension ROM. Each rolling surface, defined by radius $r_i$, forms the joint, whose center defines the joint ($J_i$) and link ($L_i$) coordinate frames. Using these parameters, the tendon hole positions can be directly obtained in both the $J_i$ and $L_i$ frames, and are incorporated into the following kinematic formulation. $R_{i}$ denotes a translation along the $x$-axis by $r_i$. Under this construction, the transformation between the link and joint frames, $T_i$, is given in (1), and the forward transformation to the next joint, $J_{i+1}$, is expressed in (2). \vspace{-0.75em}

\begin{equation}
    \begin{aligned}
        T_{i} &= 
        \begin{bmatrix}
            I_{3\times3} & t_{i} \\
            0_{1\times3} & 1
        \end{bmatrix}, \quad \text{where} \\[6pt]
        t_{i} &=
        \begin{bmatrix}
            (\kappa_{i+1}-r_{i+1}\tan(\tfrac{\beta_{i+1}}{2}))-(\kappa_{i}-r_{i}\tan(\tfrac{\beta_{i}}{2})) \\[6pt]
            0 \\
            l_{i} - r_{i} - r_{i+1}
        \end{bmatrix}
        .
    \end{aligned}
    \label{eq:Ti}
\end{equation}

\begin{equation}
    J_{i+1} = L_{i}\mathrm{Rot}_{Y}(\tfrac{\theta_{i+1}}{2}){R_{i+1}}^2\mathrm{Rot}_{Y}(\tfrac{\theta_{i+1}}{2})
    \label{eq:Ji1}
\end{equation}

Similarly, the thumb CMC and the finger MCP joints require two DOFs. The corresponding knuckle link, shown in Fig.~\ref{fig:RollingContactKinematics}(\subref{fig:RollingContactKnuckleKinematics}), is designed to be short to prevent excessive spacing between axes, and its distal end depends on the parameters of the following phalanx ($i=1$). However, $\kappa_{0}$ was intentionally designed shorter than $\kappa_{1}$ to reduce the sharp bending of the cables caused by the short link length between the radial–ulnar and flexion–extension joints. Here, $\alpha_0$ represents one-fourth of the radial–ulnar deviation ROM, and $l_{0}$ denotes the distance to the rolling surface. Accordingly, $T_0$ and $J_0$ are expressed in (\ref{eq:T0}) and (\ref{eq:J0}), respectively. \vspace{-0.5em}

\begin{equation}
    T_{0} = 
    \begin{bmatrix}
        & & & 0\\
        & I_{3\times3} & & 0\\
        & & & l_{0} - r_{1} - r_{0} \\
        & 0_{1\times3} & & 1
    \end{bmatrix}\\
    \label{eq:T0}
\end{equation}

\begin{equation}
    J_{0} = O\mathrm{Rot}_{X}(\tfrac{\phi_{0}}{2}){R_{0}}^2\mathrm{Rot}_{X}(\tfrac{\phi_{0}}{2})
    \label{eq:J0}
\end{equation}

\begin{figure}[tpb]
    \centering
    \begin{subfigure}{0.45\columnwidth}
        \centering
        \includegraphics[width=\linewidth]{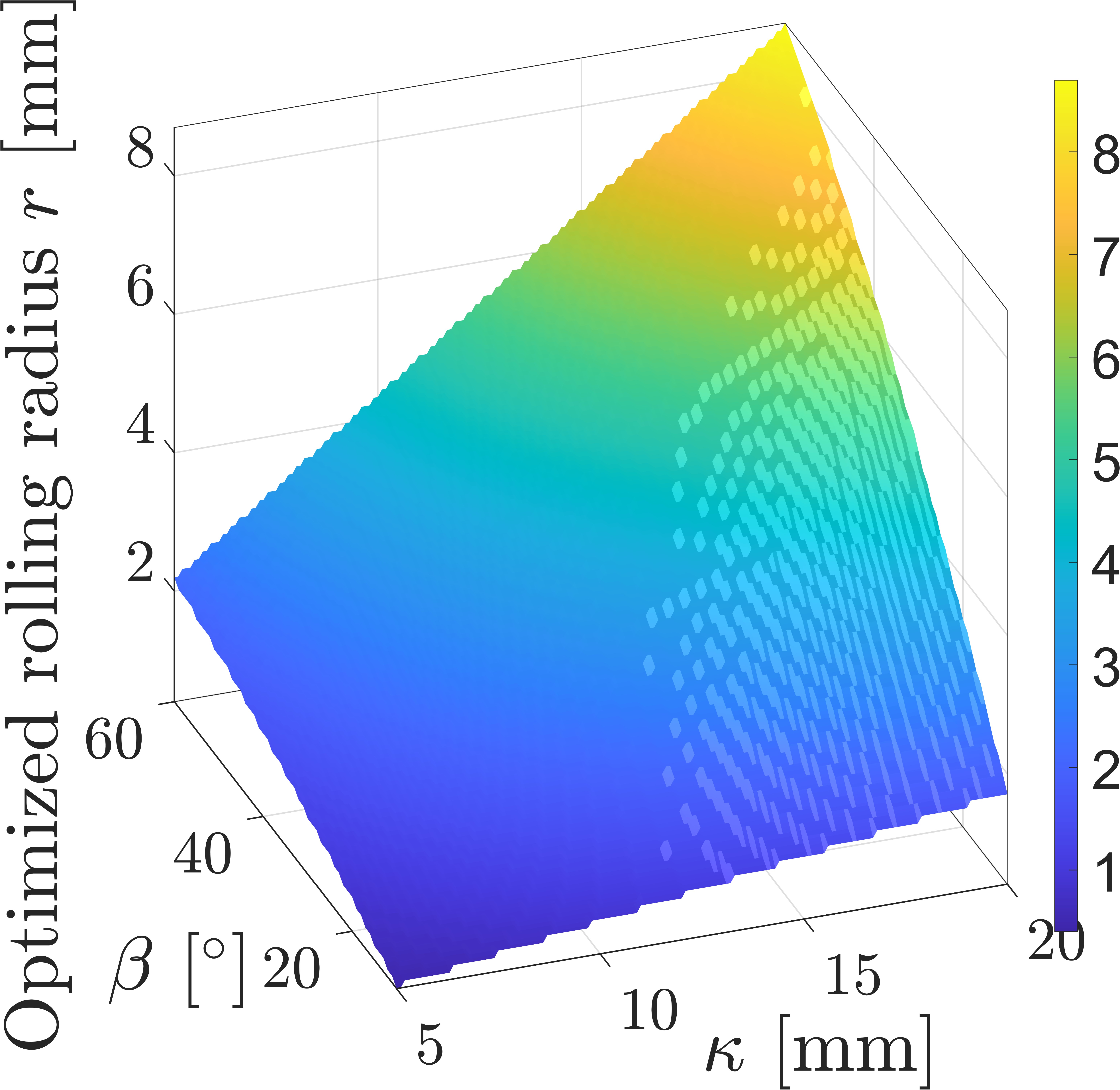}
        \caption{Optimized $r$ over ROM}
        \label{fig:RCJCableDiffSimulationA}
    \end{subfigure}
    \hfill
    \begin{subfigure}{0.49\columnwidth}
        \centering
        \includegraphics[width=\linewidth]{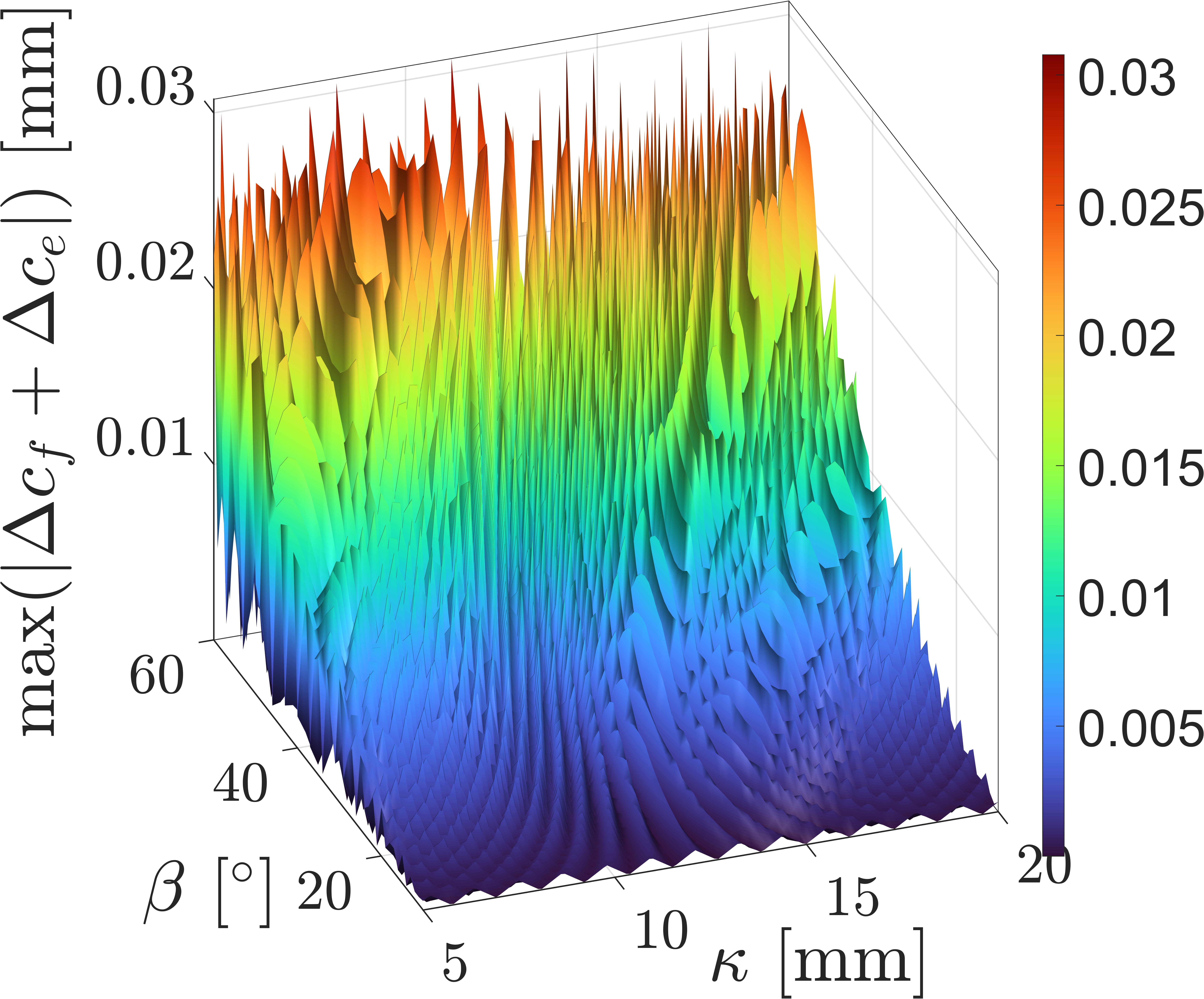}
        \caption{Cable length deviation}
        \label{fig:RCJCableDiffSimulationB}
    \end{subfigure}
    \caption{Kinematic optimization of rolling radius $r_i$ for antagonistic cable actuation. (a) Optimized $r_i$ for a flexion–extension joint, determined across different combinations of $\kappa$ and $\beta$ over the full ROM. (b) Residual maximum absolute length deviation of the antagonistic cable pair, $\max(|\Delta c_f + \Delta c_e|)$, evaluated at the optimized $r_i$ in (a).}
    \label{fig:RCJCableDiffSimulation}
    \vspace{-18pt}
\end{figure}

An important distinction is that, in the flexion–extension joint of Fig.~\ref{fig:FingerStructure}, all cables undergo identical transformations, whereas in the radial–ulnar deviation joint, only the radial and ulnar cables experience the same length change. Consequently, for single-motor antagonistic actuation, the flexor–ulnar cable must be paired with the extensor–radial cable, and the flexor–radial cable must be paired with the extensor–ulnar cable as illustrated. Here, \textit{flexors} and \textit{extensors} refer to the cables (including the radial and ulnar cables in each direction) for each RCJ. The rolling radius $r_i$ is then optimized such that the sum of length variations of the flexors ($\Delta c_f$) and extensors ($\Delta c_e$) satisfies $\lvert \Delta c_f + \Delta c_e \rvert \approx 0$ over the entire ROM of the corresponding joint, i.e., $\phi_i$ ($i=1\ldots3$) and $\theta$ ($i=0$). The optimized values of $r_i$ are determined through kinematic simulation, as shown in Fig.~\ref{fig:RCJCableDiffSimulation}. Across different combinations of $\kappa$ and $\beta$, the residual deviation of antagonistic cable lengths over the full ROM remained within approximately $0.03~\si{mm}$, which is negligible compared with the total cable length in each joint. This novel optimization scheme enables each RCJ to be driven by a single motor through antagonistic cable actuation. The overall design parameters, including link dimensions and cable hole placements, were determined to emulate the proportions and joint arrangements of the human hand, as summarized in Table~\ref{Tab:designedParameters}.

\subsection{Bowden-Cable Actuation \label{Sec:3A}}
Bowden cables are mechanisms that transmit tensile forces through a flexible inner wire supported by an outer sheath, enabling remote actuation along a curved path. This transmission method allows actuation forces to be delivered independently of the sheath geometry, as the inner cable length is preserved while the sheath resists compression. Owing to this characteristic, Bowden-cable actuation has been employed not only in robotic hands~\cite{sheath, bowdenExo} but also in surgical endoscopes~\cite{asymRCJ, invasiveRCJ}, wearable robots~\cite{wearable}, and various robotic exoskeletons~\cite{bowdenUT, bowdenArm}. Nevertheless, when the transmission path involves tight curvatures, the increased normal forces between the inner wire and the sheath result in higher friction losses~\cite{bowdenTension}. Therefore, maintaining cable routes with large bend radii is critical to improving both efficiency and service life.

\begin{figure}[tpb]
    \centering
    \includegraphics[width=0.925\columnwidth]{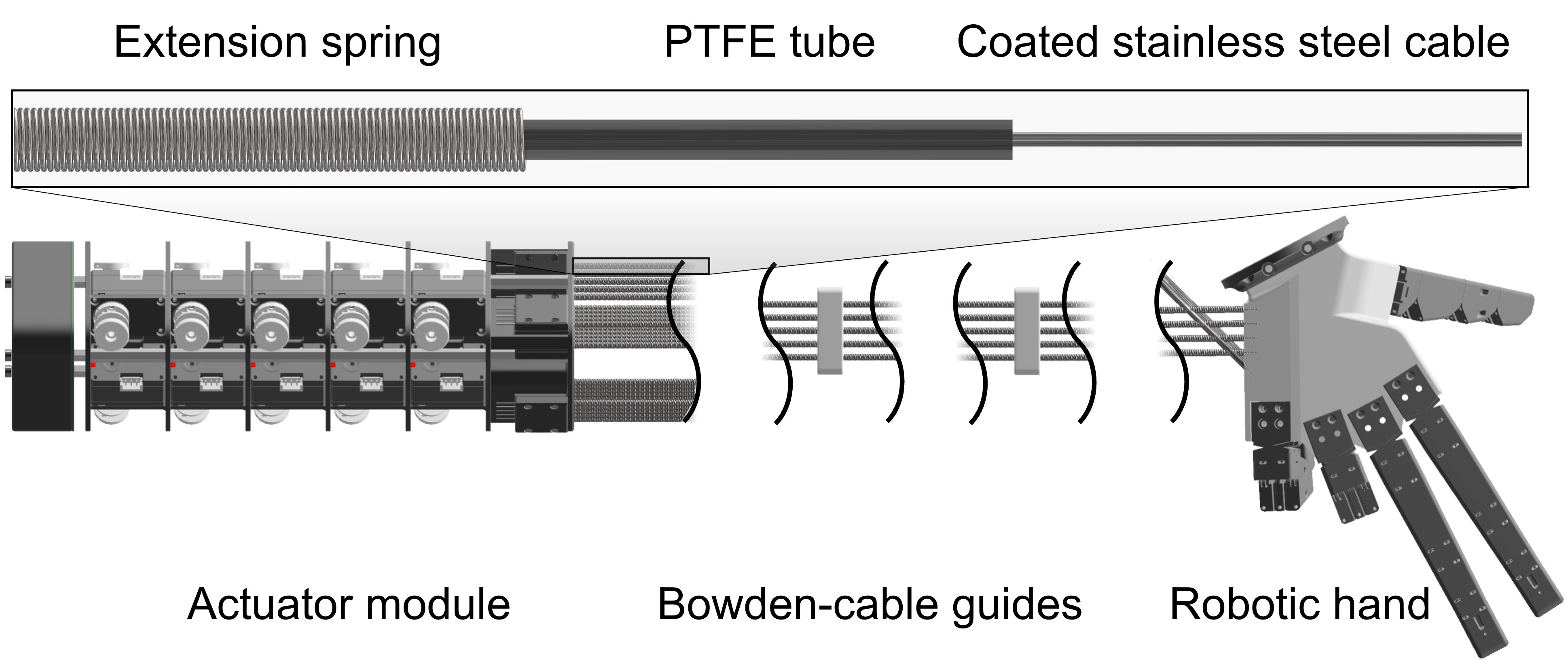}
    \caption{Bowden-cable actuation of the robotic hand and its structural components.}
    \label{fig:BowdenCable}
    \vspace{-1.5em}
\end{figure}

In this work, Bowden tubes were designed to preserve sufficient compressive stiffness to ensure stable force transmission despite long routing distances and posture variations of the robotic arm. As shown in Fig.~\ref{fig:BowdenCable}, the actuation cables employ $0.75\mathrm{mm}$ stainless-steel wires with a nylon coating, housed in polytetrafluoroethylene (PTFE) tubes with an inner diameter of $1~\mathrm{mm}$ and an outer diameter of $2~\mathrm{mm}$. To prevent collapse or damage of the PTFE sheath under cable tension, each tube was reinforced with a close-fitting tensile spring (inner diameter $2~\mathrm{mm}$, outer diameter $3.2~\mathrm{mm}$) wound around the exterior. A total of 30 Bowden-cables, forming 15 antagonistic pairs, were installed between the robotic hand and the actuator module. While the reinforcement spring itself provides some resistance to bending, additional cable guides were placed at regular intervals to further constrain excessive curvature. This arrangement prevents sharp bends, such as those occurring during elbow flexion, thereby enhancing efficiency and ensuring orderly routing of the Bowden-cables.

\subsection{Opposable Thumb and Finger Arrangement \label{Sec:3C}}

Various approaches have been explored to reproduce human thumb opposability in robotic hands. Some designs configure the thumb in a permanently opposed orientation to emphasize grasping capability~\cite{ILDA}, while others introduce non-anthropomorphic joint axes, such as roll–pitch combinations in place of the human pitch–yaw configuration, to enhance opposability~\cite{KITECH}. In addition, many hands with modular finger designed for ease of fabrication and maintenance achieve high opposability by carefully arranging the finger bases within the palm~\cite{PCDRH, ILDA, casia}.

Similarly, the palmar arches of the human hand facilitate surface contact during grasping~\cite{arches}. In robotic hand design, most efforts have concentrated on thumb opposability, and the contribution of palmar arches has received relatively less attention. Nevertheless, a prior study~\cite{synergistic} investigated the role of palm concavity and adaptability in a soft synergistic hand, showing that an articulated palm can improve contact area and grasp adaptability. Likewise, the Schunk's SVH hand incorporates a palm joint that bends inward together with the thumb to form an arch, thereby enhancing opposition and surface contact~\cite{SVH}.

\begin{figure}[tpb]
    \centering
    \includegraphics[width=0.9\columnwidth]{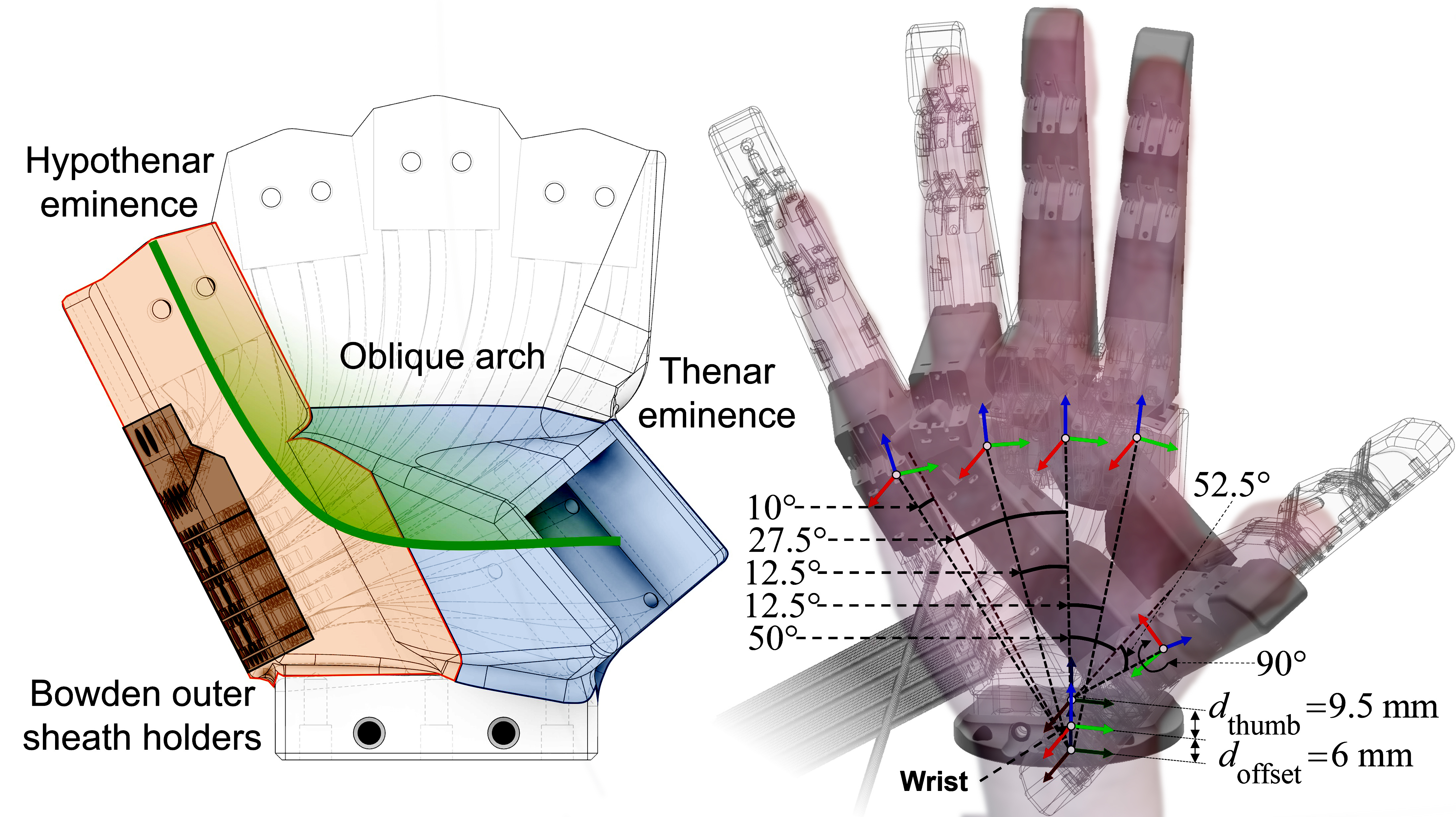}
    \caption{Palm and finger arrangement of the robotic hand with human hand overlaid for comparison of scale and placement.}
    \label{fig:palmFingerArrangement}
    \vspace{-1.5em}
\end{figure}

In the proposed robotic hand, the fingers are arranged to maintain anthropomorphic placement, dimensions, and ranges of motion while also incorporating an opposable thumb, as illustrated in Fig.~\ref{fig:palmFingerArrangement}. Human hand dimensions were measured to define the link lengths, and the relative placement of the fingers within the palm was determined by simulating full ROM postures. Transparent overlays illustrate multiple finger configurations, showing how thumb–finger interaction was considered in the design to ensure anthropomorphic opposability. Since the fingers were modularized, the transformation of finger placement within the palm became a critical design factor. In particular, we considered the oblique arch formed by the thumb and little finger as an important anatomical feature that supports thumb opposability. To mimic this function, the placement of the thumb and little finger was determined through additional transformations and analysis of inter-finger interactions, whereas the other fingers were arranged by simple radial abduction from the wrist reference point. Additionally, the base of the thumb metacarpal was designed with a broadened profile to mimic the thenar eminence, thereby enhancing the effective surface for contact during grasping.
The degree of thumb opposability achieved by this arrangement was further evaluated using a quantitative index, as described in Sec.~\ref{Sec:4A}.

\subsection{Actuator Unit \label{Sec:3D}}
The actuator module was developed using Dynamixel XC330-T288-T servomotors from Robotis Co., whose specifications are summarized in Table~\ref{tab:MotorSpec}. The Dynamixel series integrates a motor, encoder, and controller in a compact package, allowing modular arrangement depending on the application. In the proposed robotic hand, each finger requires three antagonistic cable pairs, and thus three motors per finger. For the prototype, we designed a stackable actuator module in which each layer accommodates three motors. Five layers were stacked to build the experimental module. Power and communication were routed through a dedicated bottom circuit layer. To account for the connector's current limits and communication wire's impedance characteristics, the power lines were distributed in a tree-branch configuration by layer, while the communication lines were daisy-chained sequentially through all layers.

\begin{table}[t]
  \caption{Specifications of the Motor}
  \label{tab:MotorSpec}
  \centering
  \renewcommand{\arraystretch}{1.15}
  \begin{tabular}{c | l l}
    \toprule
    \textbf{Appearance} & \textbf{Item} & \textbf{Value} \\
    \midrule
    \multirow{7}{*}{\includegraphics[width=0.20\linewidth]{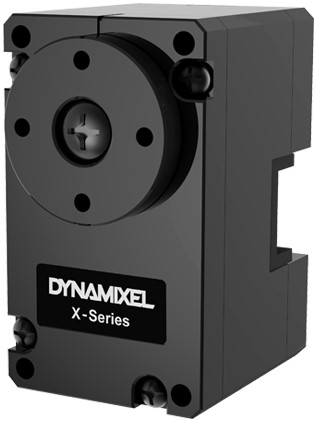}} 
    & Model name         & XC330-T288-T \\
    & Input voltage      & DC 11.1~V (typ.) \\
    & Stall torque       & 0.92~N$\cdot$m \\
    & Stall current      & 0.8~A \\
    & Encoder resolution & 4096~counts/rev \\
    & Dimensions         & $20 \times 34 \times 26~\si{mm^3}$ \\
    & Weight             & 23~g \\
    \bottomrule
  \end{tabular}
  \vspace{-0.75em}
\end{table}


An additional top layer was dedicated to bowden-cable anchoring, where the steel cables, PTFE liners, and reinforcing springs are separated and fixed. Each cable is guided downward to a bobbin mounted on the motor shaft, with antagonistic pairs wound in opposite directions. The bobbin consists of three parts: a lower barrel fixed to the motor shaft, a partition wall, and an upper barrel for securing the antagonistic pair. These parts interlock via an octagonal keyway, enabling adjustment of cable length of the antagonistic pair during assembly to set the desired preload. In the module, the flexion cables, which experience higher loads than the extension cables, were wound on the lower barrel to reduce the bending moment applied to the motor shaft.


\section{EXPERIMENTAL EVALUATION AND RESULTS \label{Sec:4}}

\begin{table}[t]
  \caption{Performance of Finger}
  \label{tab:FingerPerformance}
  \centering
  \renewcommand{\arraystretch}{1.15}
  \begin{tabular}{r | c c }
    \toprule
    Finger & Tip contact force $[\mathrm{N}]$ & Tip speed $[\mathrm{mm/s}]$\\
    \midrule
    Thumb & $21.9 \pm 0.6$ & 193 \\
    Index–Little & $18.0 \pm 0.3$ & 208 \\    
    \bottomrule
  \end{tabular}
  \vspace{-1.5em}
\end{table}

\begin{figure}[tp]
    \centering
    \begin{subfigure}{\columnwidth}
        \centering
        \includegraphics[width=0.95\linewidth]{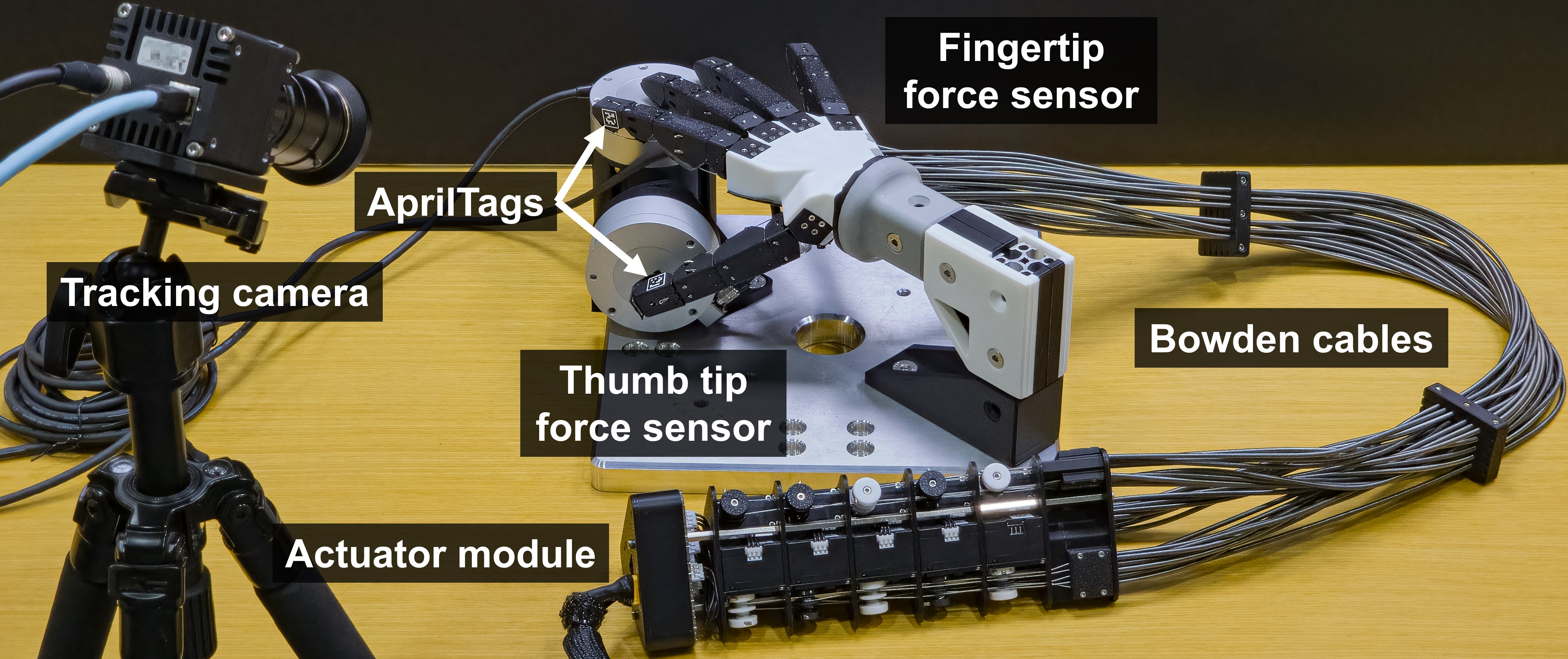}
        \caption{Experiment setup}
        \label{fig:FingertipTajStationary}
    \end{subfigure}
    \\
    \begin{subfigure}{\columnwidth}
        \centering
        \includegraphics[width=0.95\linewidth]{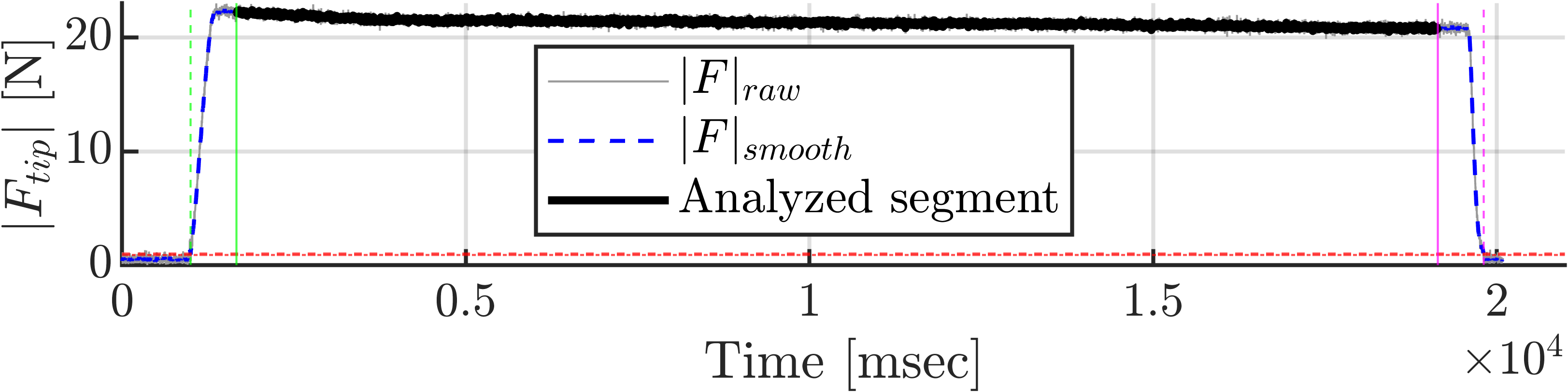}
        \caption{Measured thumb-tip contact force}
        \label{fig:FingertipTrajShaken}
    \end{subfigure}
    \caption{Finger performance measurement.}
    \label{fig:PerformanceMeasureEnvironment}
    \vspace{-1em}
\end{figure}


\subsection{Performance Evaluation \label{Sec:4A}}

The quantitative performance of the proposed hand was evaluated through measurements of fingertip contact force and tip speed, as illustrated in Fig.~\ref{fig:PerformanceMeasureEnvironment}. Fingertip force was measured by positioning a force–torque sensor near the fully extended fingertip and issuing a maximum flexion command against the sensor. Tip speed was then measured in the same setup, with the force sensor removed, by commanding maximum flexion from extension and recording the peak velocity. The results are summarized in Table~\ref{tab:FingerPerformance}.

\begin{table}[tb]
  \caption{Common Workspaces Between the Thumb and Other Fingers w.r.t. the Thumb Opposability Index}
  \label{tab:TOI}
  \centering
  \renewcommand{\arraystretch}{1.15}
  \setlength{\tabcolsep}{6pt}
  \begin{tabular}{c c c c | c}
    \toprule
    \makecell{Index\\$[\si{cm^3}]$} & \makecell{Middle\\$[\si{cm^3}]$} & \makecell{Ring\\$[\si{cm^3}]$} & \makecell{Little\\$[\si{cm^3}]$} & \makecell{Opposability Index $\mathbf{J}$} \\
    \midrule
    $88.6$ & $69.6$ & $39.6$ & $7.1$ & $0.172$ \\
    \bottomrule
  \end{tabular}
  \vspace{-1.5em}
\end{table}

\begin{figure}[tb]
    \centering
    \begin{subfigure}{0.4\columnwidth}
        \centering
        \includegraphics[width=\linewidth]{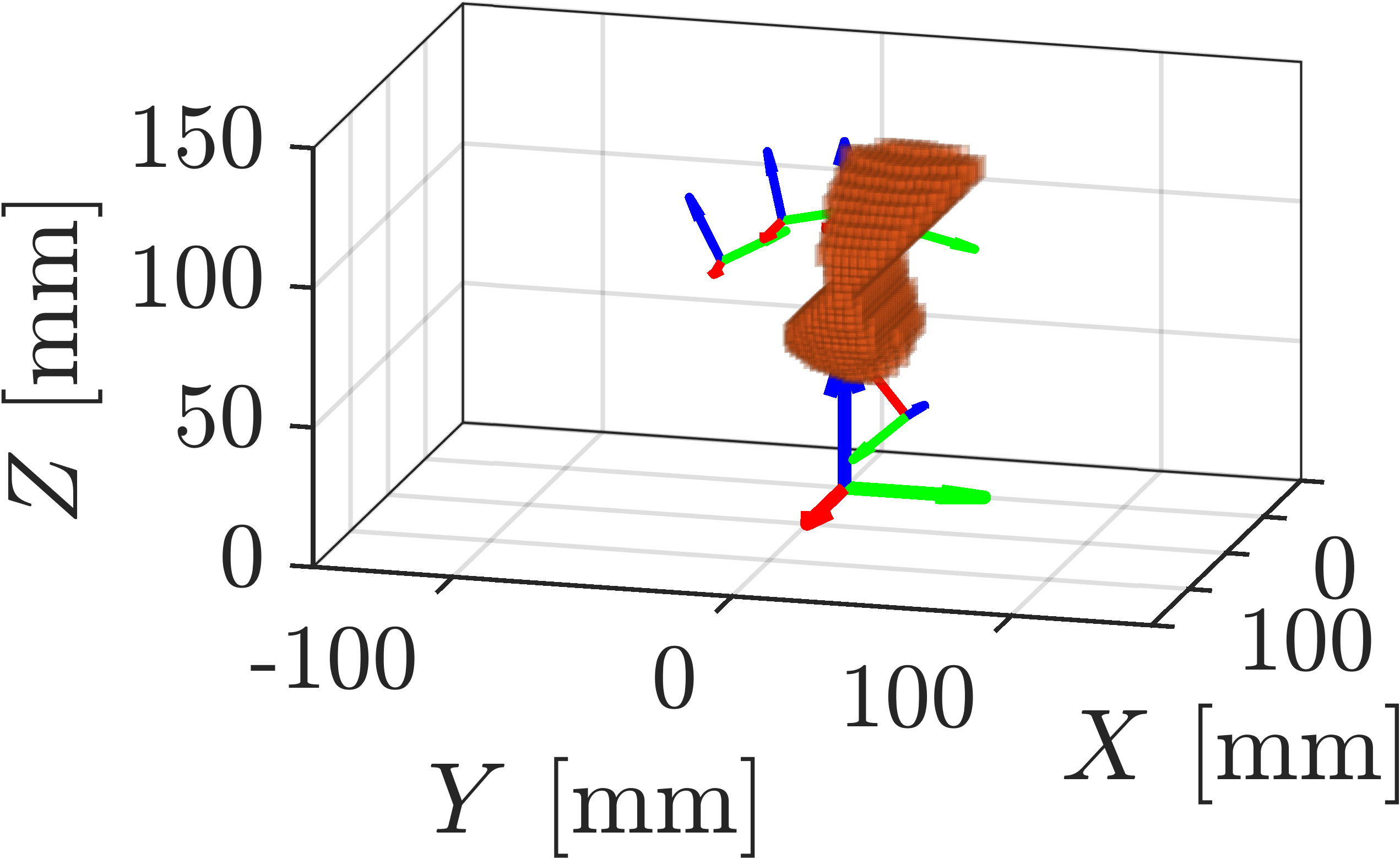}
        \caption{Thumb–Index}
    \end{subfigure}
    \begin{subfigure}{0.45\columnwidth}
        \centering
        \includegraphics[width=\linewidth]{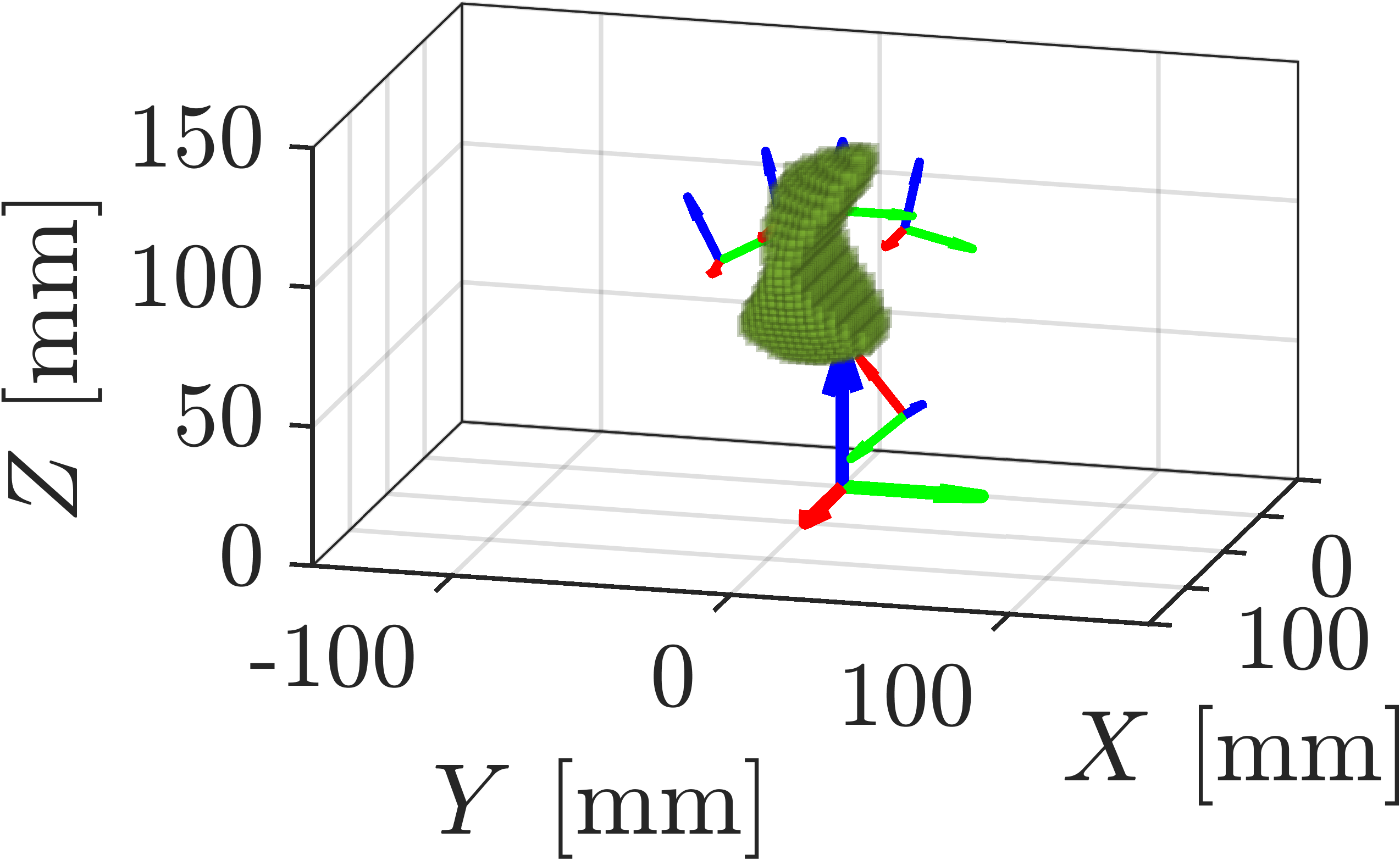}
        \caption{Thumb–Middle}
    \end{subfigure} \\
    \begin{subfigure}{0.4\columnwidth}
        \centering
        \includegraphics[width=\linewidth]{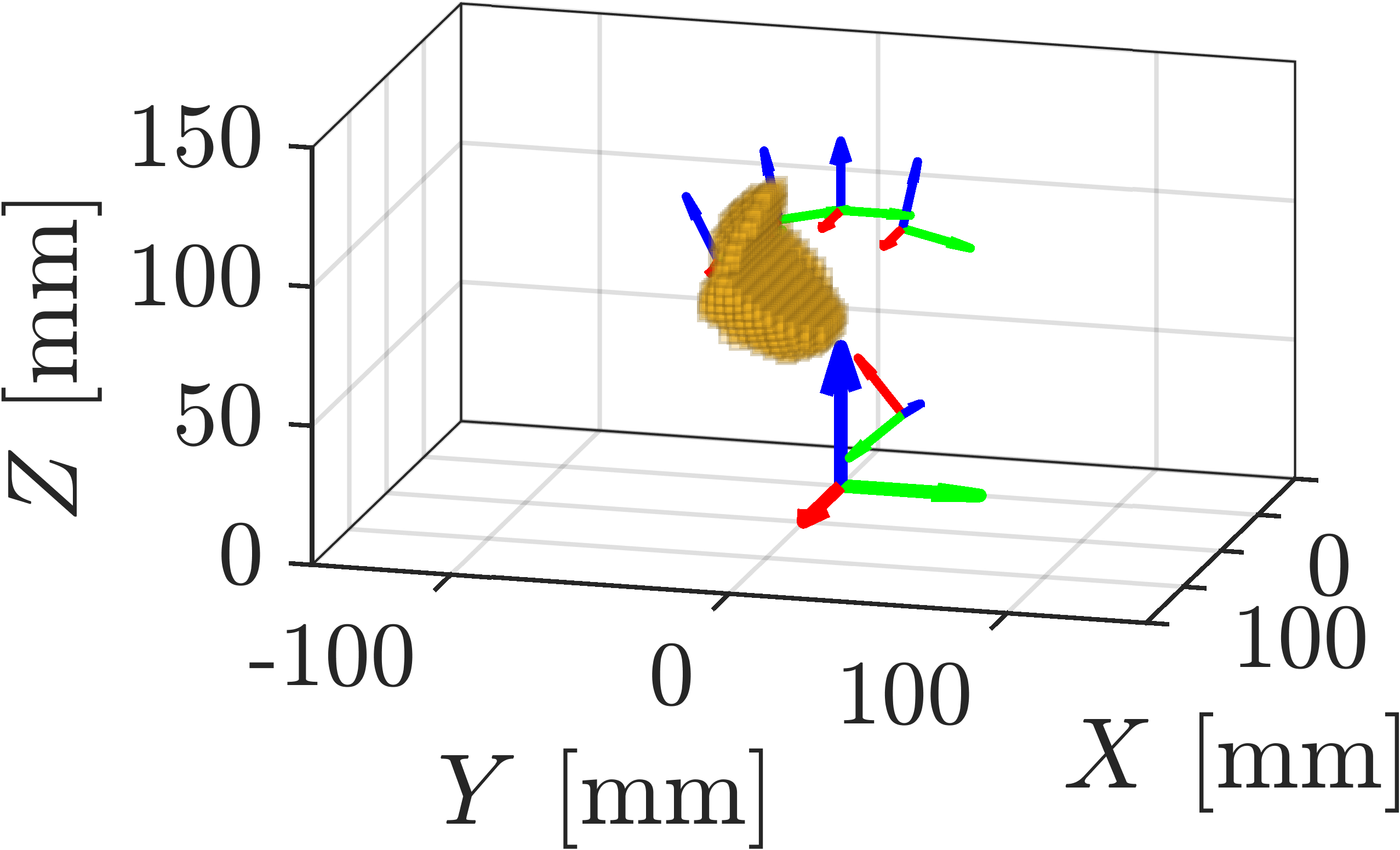}
        \caption{Thumb–Ring}
    \end{subfigure}
    \begin{subfigure}{0.4\columnwidth}
        \centering
        \includegraphics[width=\linewidth]{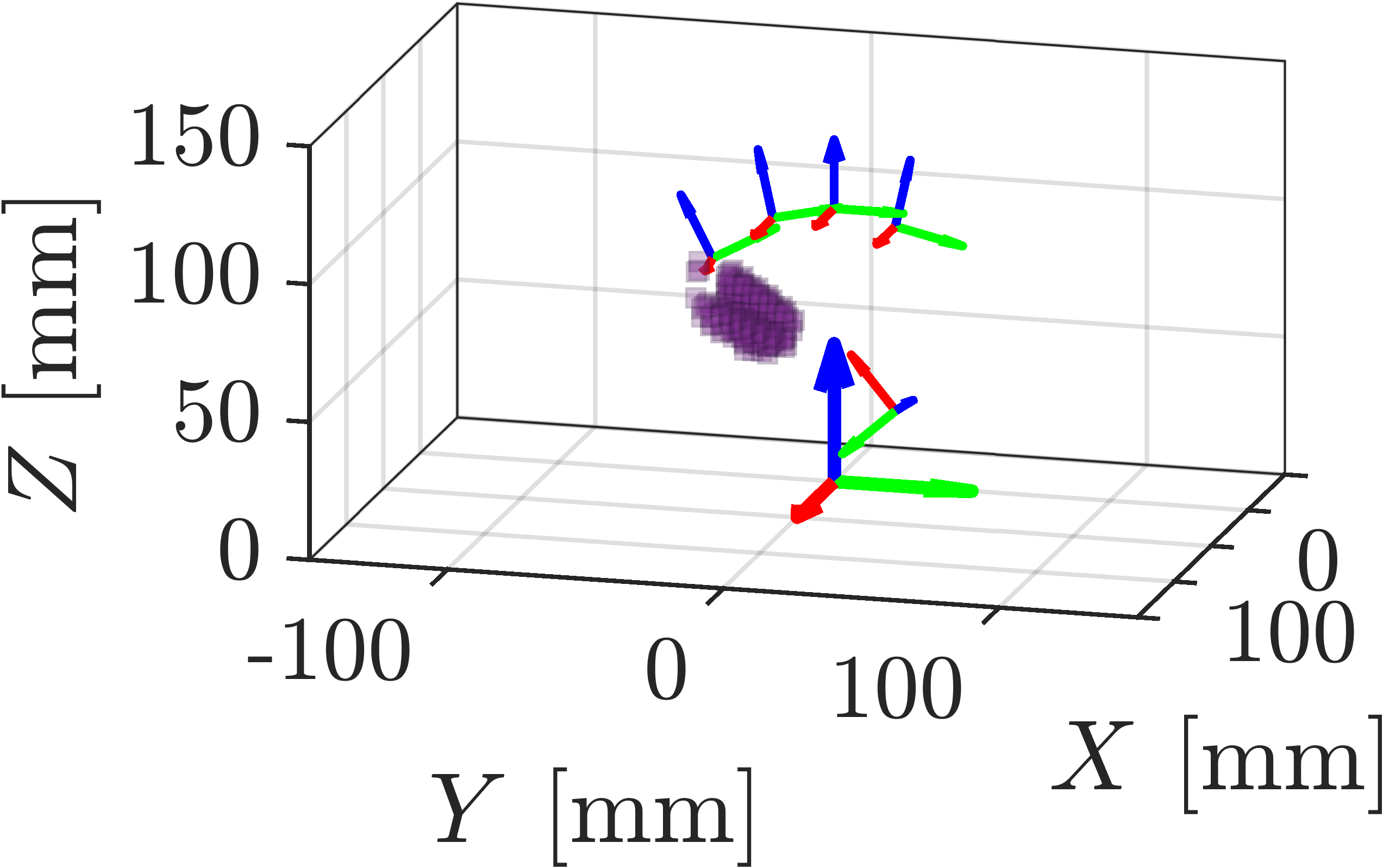}
        \caption{Thumb–Little}
    \end{subfigure} \\
    \caption{3-D representation of the common reachable workspaces between the thumb and each finger.}
    \label{fig:CommonWorkspace}
    \vspace{-1.5em}
\end{figure}

\begin{figure}[tb]
    \centering
    \begin{subfigure}{\columnwidth}
        \centering
        \includegraphics[width=\linewidth]{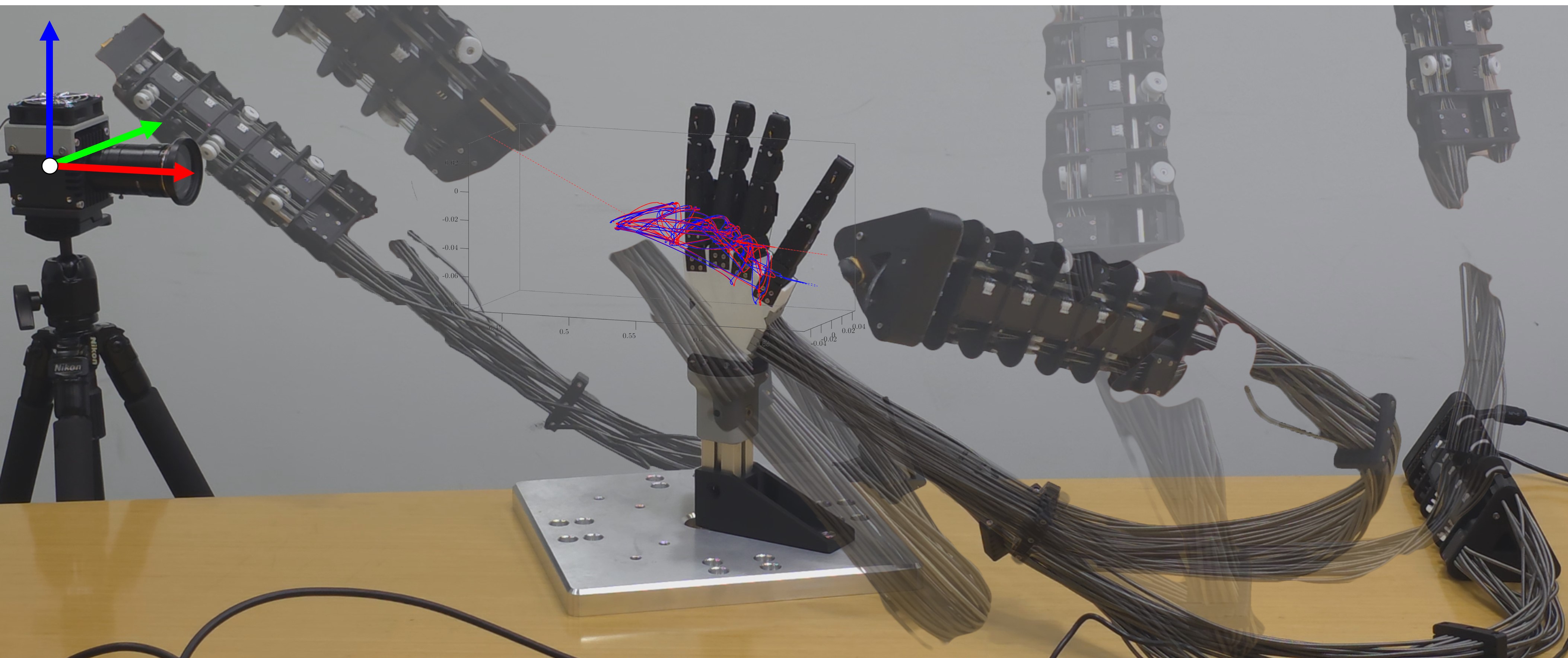}
        \caption{Actuator displacement during perturbed trial.}
        \label{fig:ActuatorMotionComposite}
    \end{subfigure}
    \\
    \begin{subfigure}{\columnwidth}
        \centering
        \includegraphics[width=0.95\linewidth]{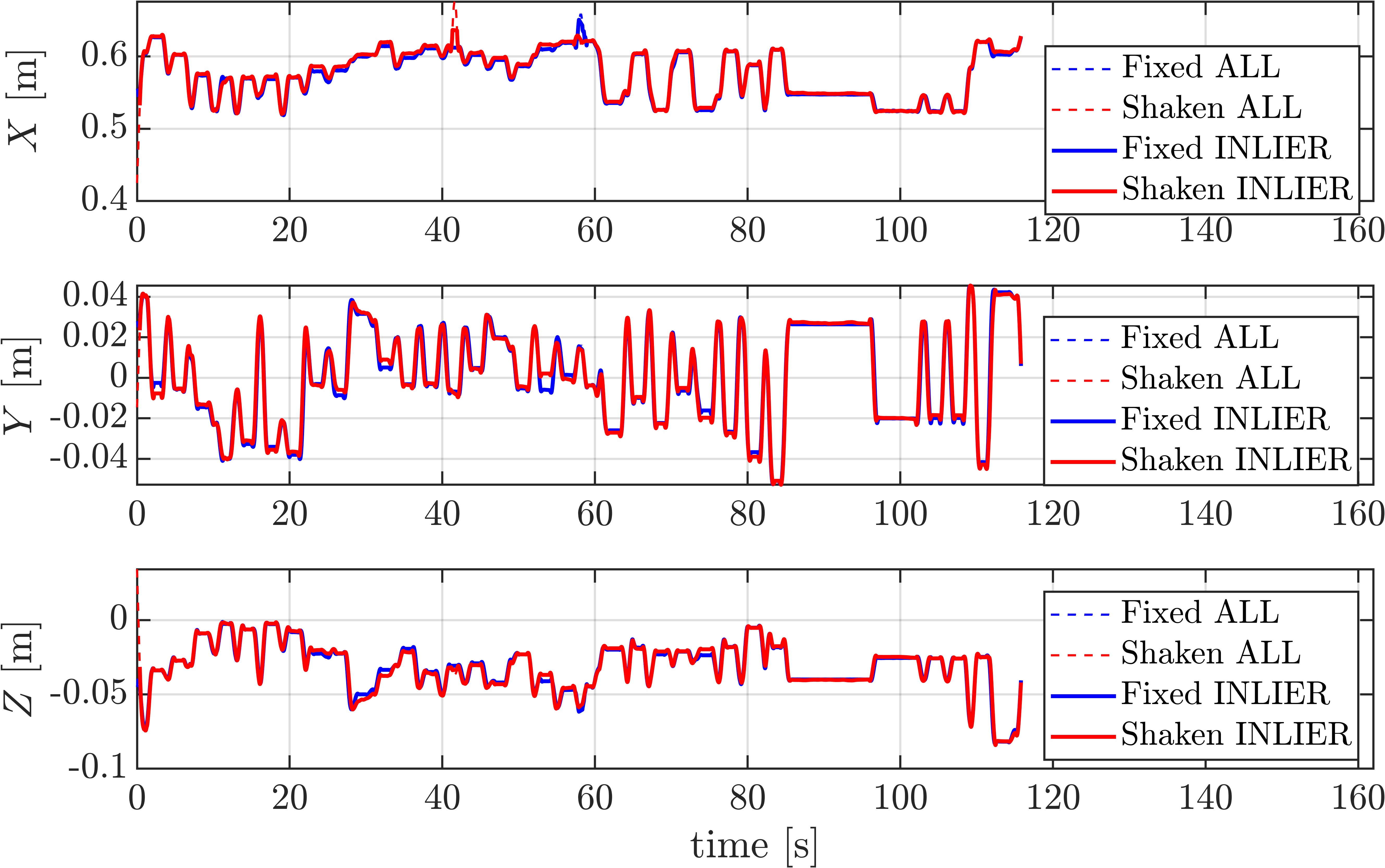}
        \caption{Overlaid fingertip trajectories (fixed vs perturbed).}
        \label{fig:FingertipTrajOverlay}
    \end{subfigure}
    \caption{Fingertip trajectory consistency with perturbed actuator-hand transformations.}
    \label{fig:BowdenRobustness}
    \vspace{-1em}
\end{figure}

\begin{figure}[tb]
    \centering
    \includegraphics[width=0.95\columnwidth]{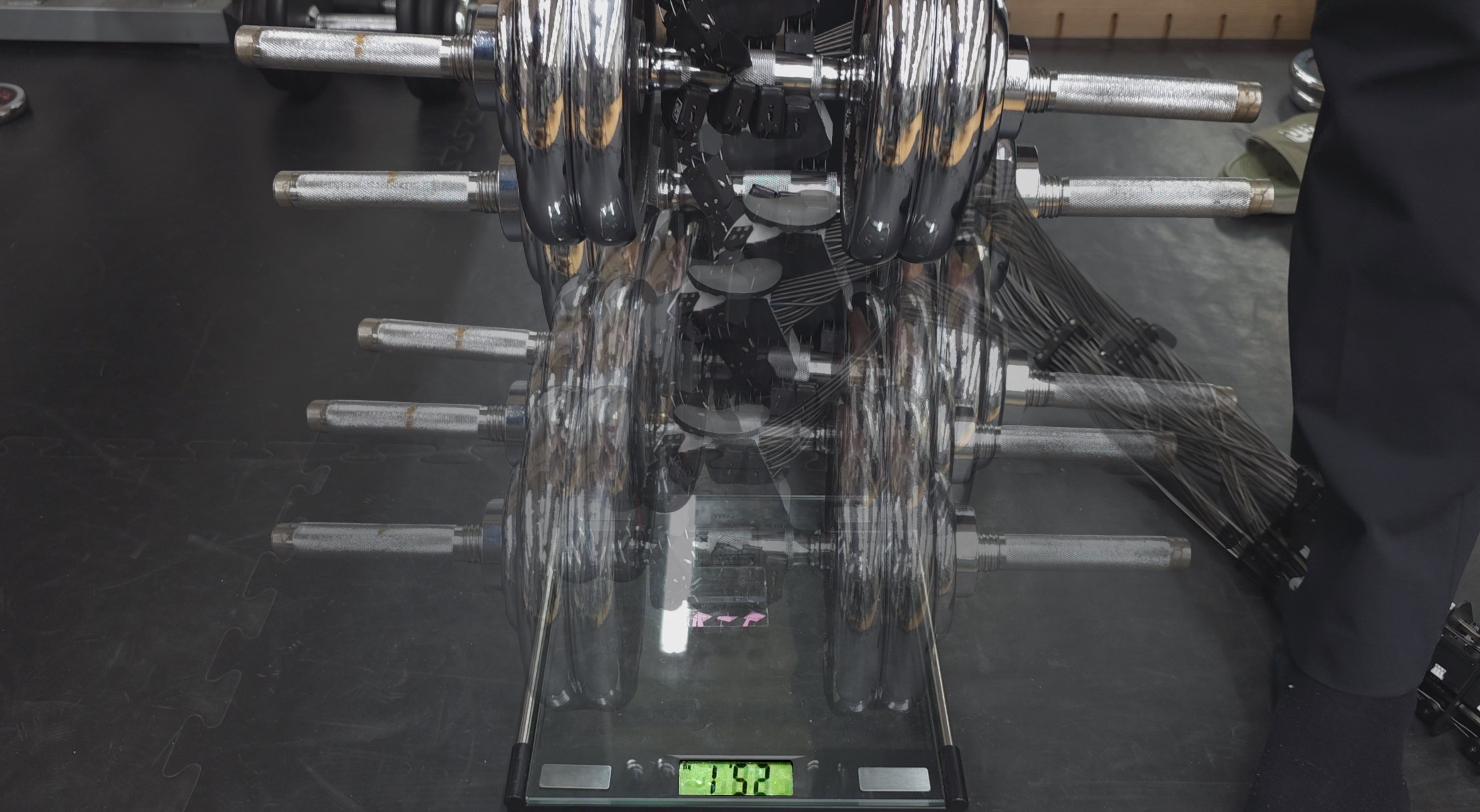}
    \caption{Power-wrap grasp lifting of a $25~\mathrm{kg}$ dumbbell.}
    \label{fig:palmFingerArrangement}
    \vspace{-1.5em}
\end{figure}

\begin{figure*}[tb]
    \centering
    \begin{subfigure}{0.118\textwidth}
        \centering
        \includegraphics[width=\linewidth]{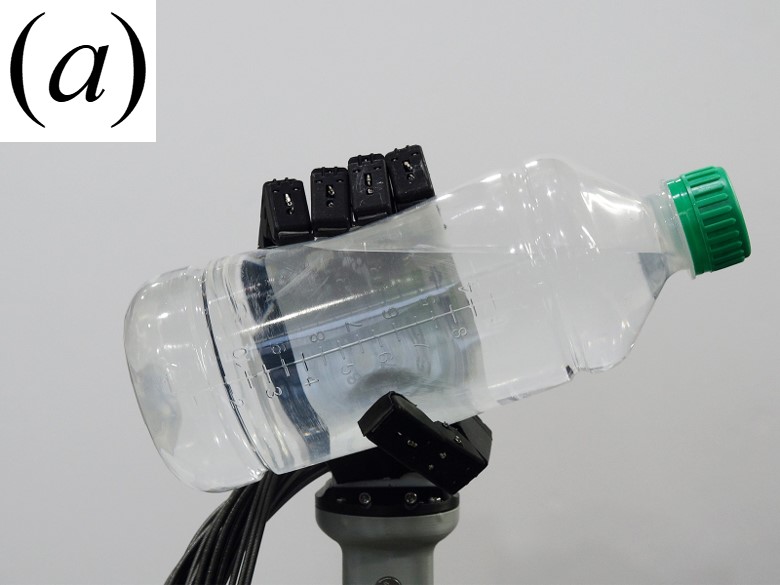}
    \end{subfigure}
    \hfill
    \begin{subfigure}{0.118\textwidth}
        \centering
        \includegraphics[width=\linewidth]{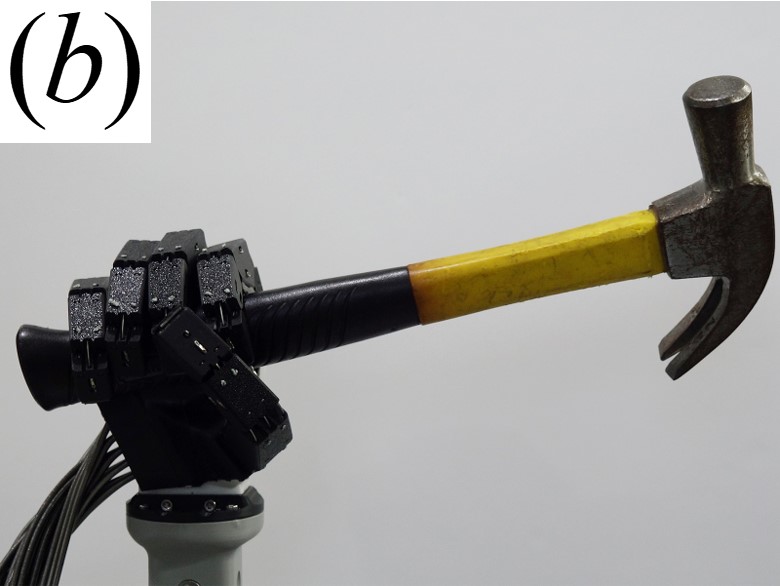}
    \end{subfigure}
    \hfill
    \begin{subfigure}{0.118\textwidth}
        \centering
        \includegraphics[width=\linewidth]{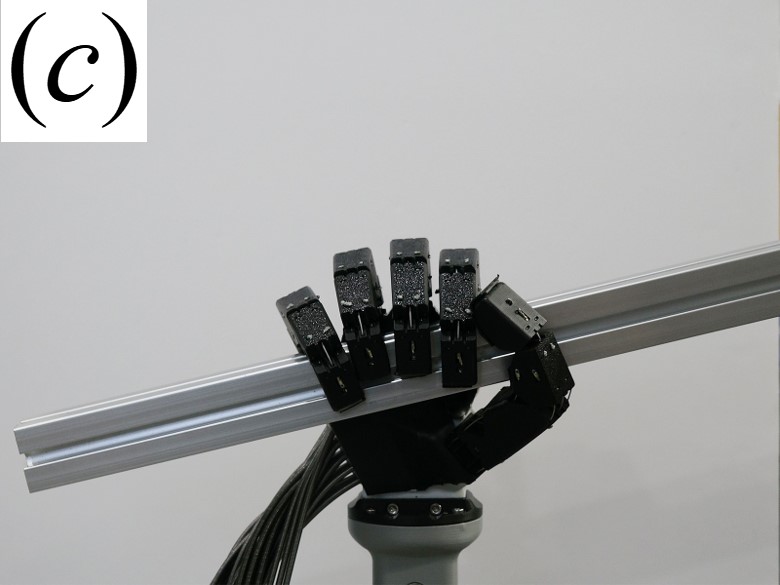}
    \end{subfigure}
    \hfill
    \begin{subfigure}{0.118\textwidth}
        \centering
        \includegraphics[width=\linewidth]{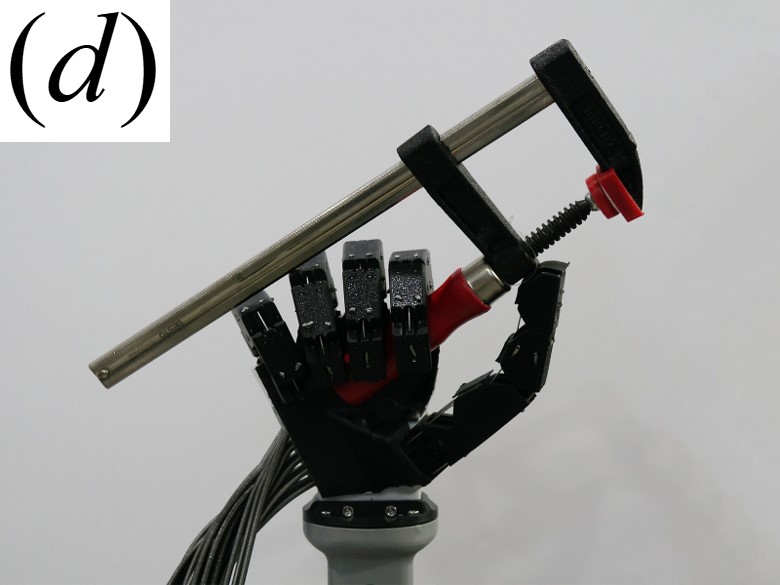}
    \end{subfigure}
    \hfill
    \begin{subfigure}{0.118\textwidth}
        \centering
        \includegraphics[width=\linewidth]{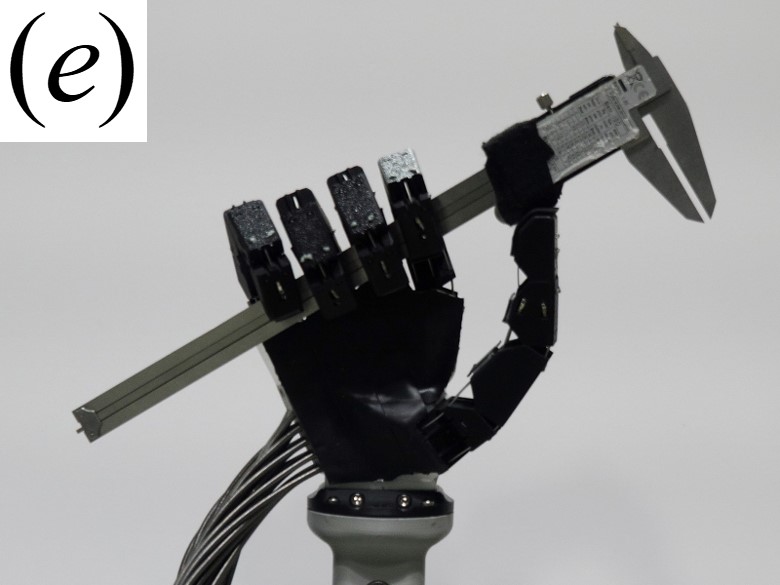}
    \end{subfigure}
    \hfill
    \begin{subfigure}{0.118\textwidth}
        \centering
        \includegraphics[width=\linewidth]{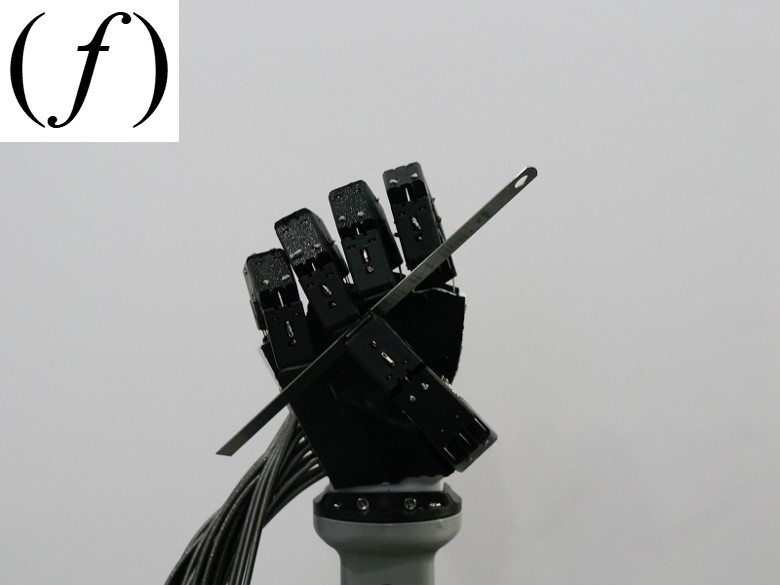}
    \end{subfigure}
    \hfill
    \begin{subfigure}{0.118\textwidth}
        \centering
        \includegraphics[width=\linewidth]{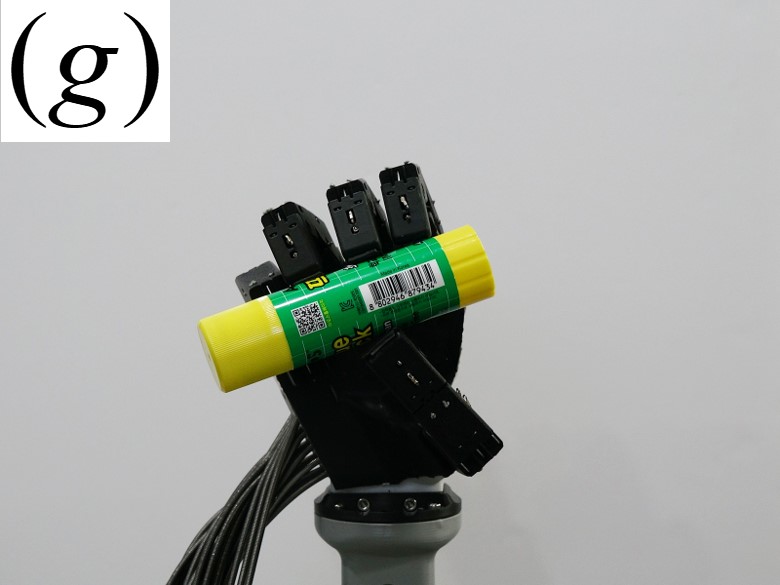}
    \end{subfigure}
    \hfill
    \begin{subfigure}{0.118\textwidth}
        \centering
        \includegraphics[width=\linewidth]{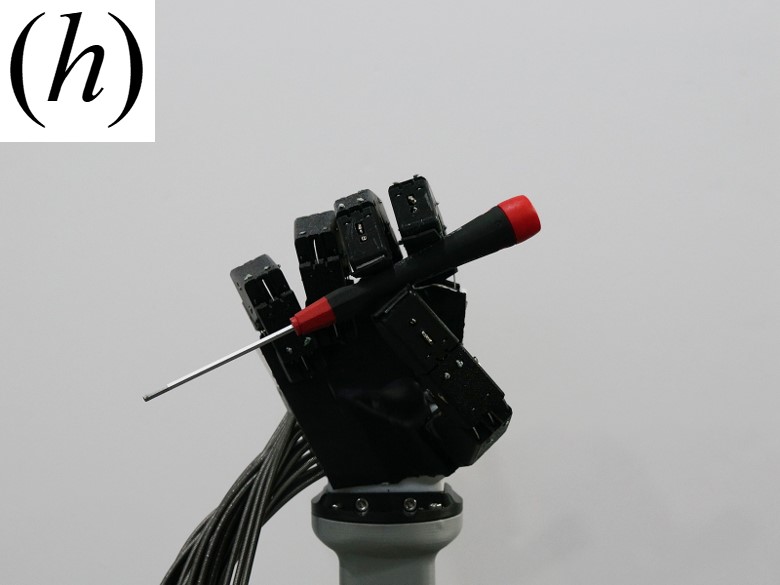}
    \end{subfigure}
    \\
    \vspace{3pt}
    \begin{subfigure}{0.118\textwidth}
        \centering
        \includegraphics[width=\linewidth]{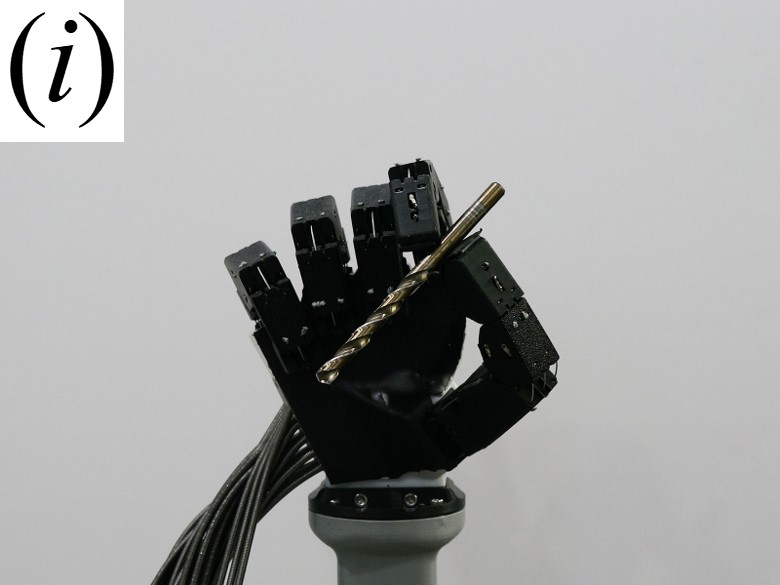}
    \end{subfigure}
    \hfill
    \begin{subfigure}{0.118\textwidth}
        \centering
        \includegraphics[width=\linewidth]{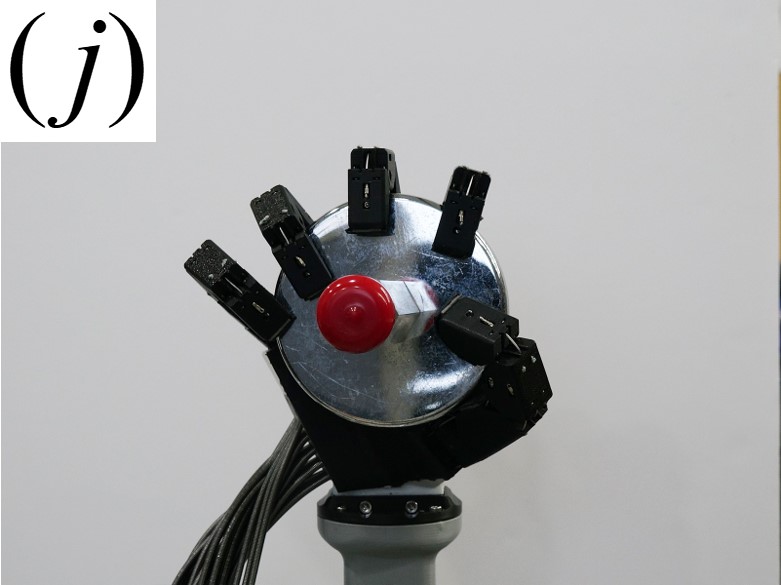}
    \end{subfigure}
    \hfill
    \begin{subfigure}{0.118\textwidth}
        \centering
        \includegraphics[width=\linewidth]{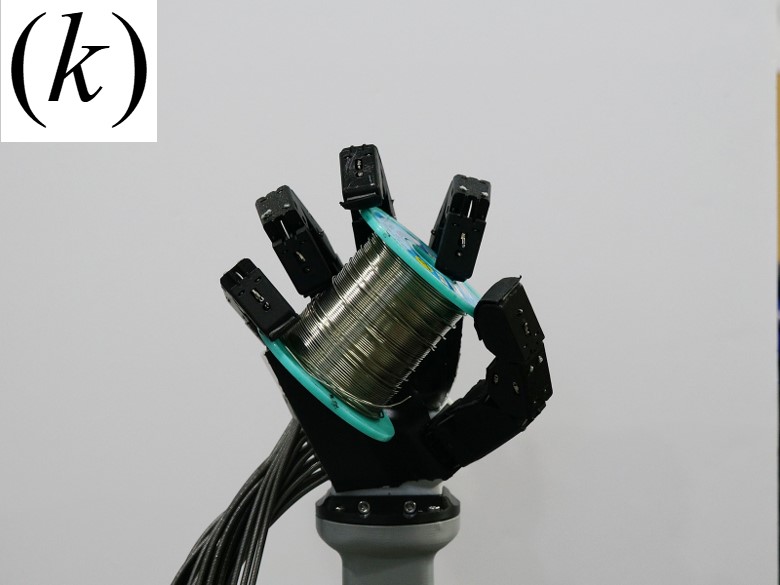}
    \end{subfigure}
    \hfill
    \begin{subfigure}{0.118\textwidth}
        \centering
        \includegraphics[width=\linewidth]{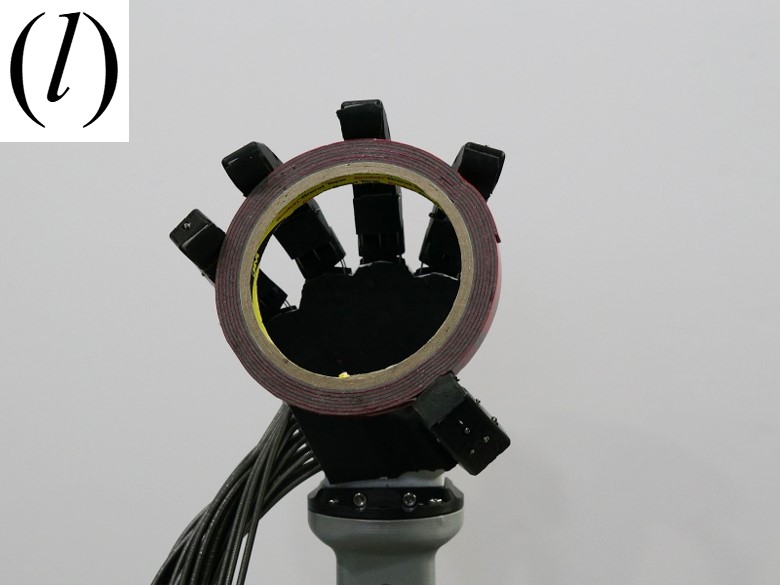}
    \end{subfigure}
    \hfill
    \begin{subfigure}{0.118\textwidth}
        \centering
        \includegraphics[width=\linewidth]{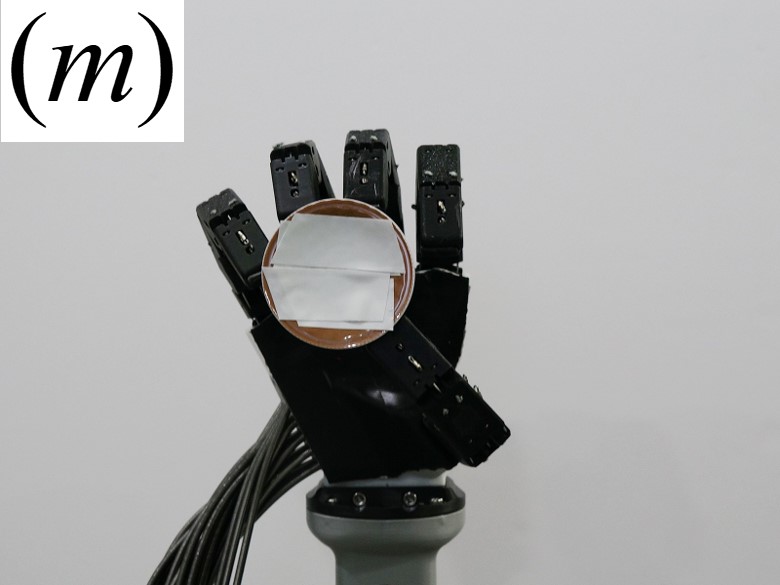}
    \end{subfigure}
    \hfill
    \begin{subfigure}{0.118\textwidth}
        \centering
        \includegraphics[width=\linewidth]{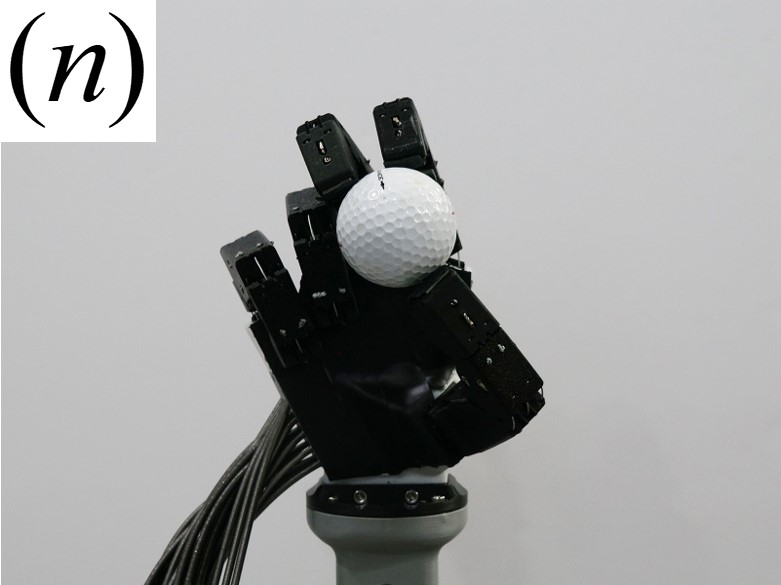}
    \end{subfigure}
    \hfill
    \begin{subfigure}{0.118\textwidth}
        \centering
        \includegraphics[width=\linewidth]{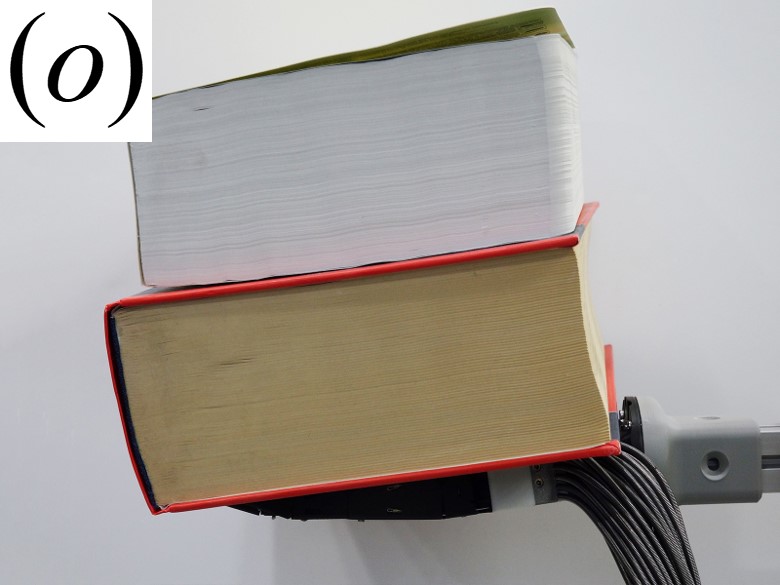}
    \end{subfigure}
    \hfill
    \begin{subfigure}{0.118\textwidth}
        \centering
        \includegraphics[width=\linewidth]{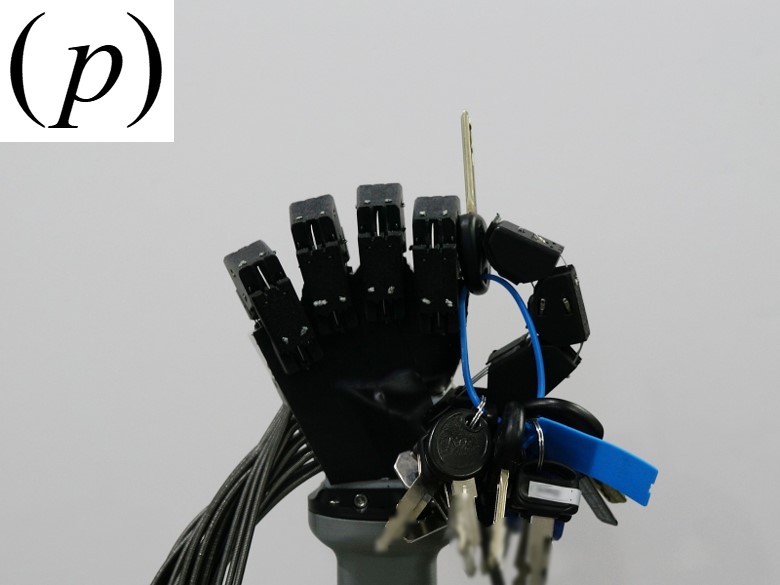}
    \end{subfigure}

    \caption{Representative grasp postures of the proposed robotic hand following Cutkosky’s taxonomy.}
    \label{Fig:CutkoskyGraspExperiment}
    \vspace{-1.25em}
\end{figure*}

\begin{table*}[tb]
    \caption{Cutkosky Grasp Taxonomy With Representative Objects and Weights, Corresponding to Fig.~\ref{Fig:CutkoskyGraspExperiment}}

    \label{Tab:CutkoskyGraspExperiment}
    \centering
    \setlength{\tabcolsep}{4pt}
  
    \newcolumntype{M}{>{\centering\arraybackslash}m{0.23\textwidth}}
    \begin{tabular}{M|M|M|M}
    \Xhline{1pt}
    \makecell[c]{\scriptsize Fig.~\ref{Fig:CutkoskyGraspExperiment}(a): \textbf{Large heavy wrap}\\
                 \scriptsize Object: Bottle (\SI{1.075}{\kilogram})}
    &
    \makecell[c]{\scriptsize Fig.~\ref{Fig:CutkoskyGraspExperiment}(b): \textbf{Small heavy wrap}\\
                 \scriptsize Object: Hammer (\SI{0.800}{\kilogram})}
    &
    \makecell[c]{\scriptsize Fig.~\ref{Fig:CutkoskyGraspExperiment}(c): \textbf{Medium wrap}\\
                 \scriptsize Object: Aluminum profile (\SI{0.395}{\kilogram})}
    &
    \makecell[c]{\scriptsize Fig.~\ref{Fig:CutkoskyGraspExperiment}(d): \textbf{Adducted thumb}\\
                 \scriptsize Object: Clamp (\SI{0.375}{\kilogram})}
    \\
    \hline
    \makecell[c]{\scriptsize Fig.~\ref{Fig:CutkoskyGraspExperiment}(e): \textbf{Light tool}\\
                 \scriptsize Object: Calipers (\SI{0.195}{\kilogram})}
    &
    \makecell[c]{\scriptsize Fig.~\ref{Fig:CutkoskyGraspExperiment}(f): \textbf{Thumb--4 fingers}\\
                 \scriptsize Object: Ruler (\SI{0.012}{\kilogram})}
    &
    \makecell[c]{\scriptsize Fig.~\ref{Fig:CutkoskyGraspExperiment}(g): \textbf{Thumb--3 fingers}\\
                 \scriptsize Object: Stick glue (\SI{0.062}{\kilogram})}
    &
    \makecell[c]{\scriptsize Fig.~\ref{Fig:CutkoskyGraspExperiment}(h): \textbf{Thumb--2 fingers}\\
                 \scriptsize Object: Hex driver (\SI{0.019}{\kilogram})}
    \\
    \hline
    \makecell[c]{\scriptsize Fig.~\ref{Fig:CutkoskyGraspExperiment}(i): \textbf{Thumb--Index finger}\\
                 \scriptsize Object: Drill bit (\SI{0.019}{\kilogram})}
    &
    \makecell[c]{\scriptsize Fig.~\ref{Fig:CutkoskyGraspExperiment}(j): \textbf{Power disk}\\
                 \scriptsize Object: Magnet handle (\SI{0.830}{\kilogram})}
    &
    \makecell[c]{\scriptsize Fig.~\ref{Fig:CutkoskyGraspExperiment}(k): \textbf{Power sphere}\\
                 \scriptsize Object: Solder (\SI{0.465}{\kilogram})}
    &
    \makecell[c]{\scriptsize Fig.~\ref{Fig:CutkoskyGraspExperiment}(l): \textbf{Precision disk}\\
                 \scriptsize Object: Double-sided tape (\SI{0.030}{\kilogram})}
    \\
    \hline
    \makecell[c]{\scriptsize Fig.~\ref{Fig:CutkoskyGraspExperiment}(m): \textbf{Precision sphere}\\
                 \scriptsize Object: Coffee capsule (\SI{0.025}{\kilogram})}
    &
    \makecell[c]{\scriptsize Fig.~\ref{Fig:CutkoskyGraspExperiment}(n): \textbf{Tripod}\\
                 \scriptsize Object: Golf ball (\SI{0.046}{\kilogram})}
    &
    \makecell[c]{\scriptsize Fig.~\ref{Fig:CutkoskyGraspExperiment}(o): \textbf{Platform push}\\
                 \scriptsize Object: Books (\SI{7.560}{\kilogram})}
    &
    \makecell[c]{\scriptsize Fig.~\ref{Fig:CutkoskyGraspExperiment}(p): \textbf{Lateral pinch}\\
                 \scriptsize Object: Key ring (\SI{0.137}{\kilogram})}
    \\
    \Xhline{1pt}
  \end{tabular}
  \vspace{-1em}
\end{table*}

In addition to force and speed measurements, the widely adopted thumb opposability index ($\mathbf{J}$)~\cite{gifuIII} was used to evaluate the proposed design. The opposability index is defined as
\begin{equation}
    \mathbf{J} = \frac{1}{d^3} \sum^{k}_{i=1} w_i v_{i},
    \label{eq:ThumbOpposabilityIndex}
\end{equation}
where $d$ denotes the thumb length, $w_i$ is the weight factor for each finger (set to $1$ for all fingers in this letter), and $v_i$ represents the shared workspace between the thumb and the $i$-th finger. This index has been widely adopted as a quantitative benchmark to assess thumb–finger coordination in diverse anthropomorphic robotic hand designs~\cite{gifuIII, KITECH, PCDRH}. It should be noted that in this work the numerical simulation of opposability was not employed to optimize finger placement, but rather to objectively quantify the extent to which the proposed design preserves opposability. As summarized in Table~\ref{tab:TOI}, the proposed hand achieved a thumb opposability index higher than those reported in prior anthropomorphic designs.


\subsection{Antagonistic Bowden-cable Actuation \label{Sec:4B}}

To evaluate the consistency and robustness of the antagonistic Bowden-cable actuation mechanism, fingertip trajectories were compared under two conditions: a static configuration with the actuator module fixed, and a perturbed configuration where the actuator was intentionally shaken relative to the hand. This setup replicates a humanoid scenario in which actuators are mounted remotely on the torso while the hand experiences motion at the wrist. During the trials, the hand repeatedly executed all classes of the Cutkosky taxonomy along with additional dexterous postures such as rock–paper–scissors gestures, thumb-up, and thumb–finger opposition. Fingertip positions were tracked with AprilTag markers and measured by an external vision system. The trajectories from both conditions were temporally aligned, smoothed, and compared using root-mean-square error (RMSE) analysis. 

As illustrated in Fig.~\ref{fig:BowdenRobustness}(a), the actuator module experienced visible perturbations, while Fig.~\ref{fig:BowdenRobustness}(b) shows that fingertip trajectories from both conditions remained closely aligned. After processing, the RMSE of fingertip positions was about $5~\si{mm}$. This deviation reflects both the imposed actuator perturbations and the inherent variability of the underactuated cable-driven mechanism, where identical commands can yield slightly different fingertip paths. These results indicate that the antagonistic Bowden-cable transmission maintained consistent kinematic behavior within a bounded deviation under dynamic actuator–hand transformations.

\subsection{Object Manipulation \label{Sec:4C}}


The extreme payload capacity was evaluated by executing a power-wrap grasp on a $25~\mathrm{kg}$ dumbbell—over one hundred times the hand's own mass. The hand successfully lifted and held the object in 10 consecutive trials without failure, demonstrating the robustness of the Bowden-cable transmission (Fig.~\ref{fig:palmFingerArrangement}). In addition, we evaluated object grasp performance and the effectiveness of the anthropomorphic design with an opposable thumb by reproducing all grasp classes of the Cutkosky taxonomy~\cite{cutkosky}. As illustrated in Fig.~\ref{Fig:CutkoskyGraspExperiment}, the hand achieved natural grasp postures and stably held the target objects, as summarized in Table~\ref{Tab:CutkoskyGraspExperiment}. A supplementary video demonstrating the grasp experiments is available at \url{https://youtu.be/bZ6SG_xJHlY}.

\section{CONCLUSIONS \label{Sec:5}}


This letter presented the design and implementation of a lightweight anthropomorphic robotic hand featuring optimized rolling-contact joints that enable single-motor antagonistic actuation via Bowden cables. By relocating the actuator module to the torso through Bowden-cable transmission, the design substantially mitigates the distal mass limitations of prior anthropomorphic hands while maintaining human-scale dimensions and dexterous multi-DOF motion. In contrast, many prior tendon-driven or linkage-based designs concentrate actuators within the palm, increasing distal mass and reducing payload efficiency for humanoid arms. Experimental validations confirmed that the hand achieves fingertip forces sufficient for stable grasping, preserves trajectory consistency under actuator perturbations, and demonstrates robustness across diverse grasp taxonomies. These results demonstrate the feasibility of antagonistic Bowden-cable actuation as a practical solution for payload-constrained humanoids.

Future work will focus on further enhancing thumb fidelity and inter-finger coordination. In particular, we plan to investigate rolling-contact geometries at the carpometacarpal joint for ulnar–radial motion with passive roll, as well as antagonistic routing schemes inspired by intrinsic muscles such as the first dorsal interosseous. Such extensions could further enhance anthropomorphic fidelity and expand the manipulation capabilities of the proposed hand.

\addtolength{\textheight}{-12cm}   






\end{document}